\documentclass[journal]{IEEEtran}
\ifCLASSINFOpdf
\else
\fi
\usepackage{graphicx}
\usepackage{booktabs}
\usepackage{amsmath}
\usepackage{multirow}
\usepackage{graphicx}
\usepackage{multicol}
\usepackage{colortbl}
\usepackage{amssymb}

\usepackage[table]{xcolor}
\usepackage{subfigure}
\usepackage{overpic}
\usepackage{hyperref}
\usepackage[capitalize]{cleveref}

\usepackage{xcolor}
\usepackage{pgfplots}
\usepgfplotslibrary{groupplots}
\pgfplotsset{compat=1.18}
\usetikzlibrary{patterns}
\usepackage{tikz}
\definecolor{myblue}{HTML}{92dcec}

\definecolor{hous}{HTML}{b88b4d}
\definecolor{green}{HTML}{79c561}
\definecolor{farming}{HTML}{ded94c}
\definecolor{trans}{HTML}{b4b4a9}
\definecolor{services}{HTML}{ff362e}
\definecolor{other}{HTML}{dbd4d3}
\definecolor{industry}{HTML}{db79c0}
\definecolor{water}{HTML}{7982db}
\definecolor{techinfra}{HTML}{303355}

\usepackage{xcolor}
\definecolor{iccvblue}{rgb}{0.21,0.49,0.74}
\definecolor{upcolor}{rgb}{0.73, 0.15, 0.12}
\definecolor{downcolor}{rgb}{0.02,0.24,0.74}
\usepackage{lineno,hyperref}

\usepackage{bm} 
\usepackage{textcomp} 
\usepackage{multirow}
\usepackage{colortbl}
\usepackage[table]{xcolor}
\usepackage{colortbl}
\usepackage{amsmath}
\usepackage{mathrsfs}
\usepackage{booktabs}
\usepackage{float}
\makeatletter
\newcommand{\ssymbol}[1]{$^{\@fnsymbol{#1}}$}
\makeatother
\newcommand{\myparagraph}[1]{\textbf{#1}\hspace{1.8ex}}

\definecolor{aliceblue}{rgb}{0.87, 0.92, 0.96}

\usepackage{amsmath} 
\usepackage{color}
\usepackage{adjustbox}
\usepackage{url}
\usepackage{bbding}
\usepackage{amssymb}
\usepackage{bm}  
\usepackage{graphicx}  
\usepackage[ruled,vlined]{algorithm2e}
\usepackage{pifont}
\usepackage{makecell}

\crefname{section}{Sec.}{Secs.}
\Crefname{section}{Section}{Sections}
\Crefname{table}{Table}{Tables}
\crefname{table}{Tab.}{Tabs.}

\graphicspath{{figures/}}

\begin{document}
%
\title{MedVeriSeg: Teaching LISA-Like Medical Segmentation Models to Verify Query Validity Without Extra Training}
%
%
%

\author{Qinyue~Tong,
        Xiaozhen~Wang,
        Ziqian~Lu,
        Jun~Liu,
        Yunlong~Yu,~\IEEEmembership{Member,~IEEE,}
        Zheming~Lu*,~\IEEEmembership{Senior Member,~IEEE}
        
\thanks{Qinyue Tong, Jun Liu and Zheming Lu are with the School of Aeronautics and Astronautics, Zhejiang University, Hangzhou 310027, China (e-mail: \{qinyuetong, junliu0930, zheminglu\}@zju.edu.cn).}
\thanks{Xiaozhen Wang is with Southern Medical University, Guangzhou 510515, Guangdong, China (e-mail: xiaozhenwang@i.smu.edu.cn).}
\thanks{Ziqian Lu is with the School of Computer Science and Technology (School of Artificial Intelligence), Zhejiang Sci-Tech University, Hangzhou 310018, Zhejiang, China (e-mail: ziqianlu@zstu.edu.cn).}
\thanks{Yunlong Yu is with the College of Information Science and Electronic Engineering, Zhejiang University, Hangzhou, 310027 China (e-mail: yuyunlong@zju.edu.cn).}
\thanks{*Corresponding author: Zheming Lu (e-mail: zheminglu@zju.edu.cn)}
\thanks{This work has been submitted to the IEEE for possible publication. Copyright may be transferred without notice, after which this version may no longer be accessible.}
}

\maketitle

\begin{abstract}
Despite recent progress in text-prompt-based medical image segmentation, existing LISA-like MLLM-based methods typically generate masks regardless of whether the target specified in the query is present, leading to hallucinated segmentation.
In this work, we propose \textbf{MedVeriSeg}, a training-free query verification framework that enables LISA-like medical segmentation models to reject false segmentation queries.
MedVeriSeg first quantifies the response quality between the \texttt{[SEG]} token and image features through a \textbf{Similarity Response Quality Scoring Module}. To further improve robustness, it employs a \textbf{Lightweight Routed Multi-Agent Verification Module}, which fuses quantitative score evidence with qualitative agent evidence to comprehensively verify the validity of the query.
To support systematic evaluation, we construct \textbf{MedVeriSeg-Bench}, a benchmark designed for query verification in medical image segmentation.
Experimental results demonstrate that MedVeriSeg effectively identifies false segmentation queries and reduces hallucinated segmentation, while maintaining a high acceptance rate for valid queries, thereby largely preserving the segmentation utility of LISA-like medical segmentation models.
The project code has been released at \href{https://github.com/Edisonhimself/MedVeriSeg}{\textcolor{cyan}{\textit{https://github.com/Edisonhimself/MedVeriSeg}}}.
\end{abstract}



%
\IEEEpeerreviewmaketitle

\section{Introduction}\label{sec:intro}

\IEEEPARstart{M}{edical} image segmentation aims to accurately delineate regions of interest across diverse imaging modalities \cite{instruction-med-seg-onesentence-1,intro_first_seg-2, intro_first_seg-3,csvt_medical_seg_1,csvt_medical_seg_2,csvt_medical_seg_3,csvt_medical_seg_4}. 
Among various approaches, text-prompt-based methods employ textual prompts as instructions to guide models in segmenting target regions within medical images \cite{biomed-parse,imis_model,yan2025medreasoner}. 
This paradigm substantially enhances the interactivity of medical image segmentation by enabling more flexible and intuitive human–model interactions across diverse settings.

\begin{figure}[t]
    \centering
    \includegraphics[width=1\linewidth]{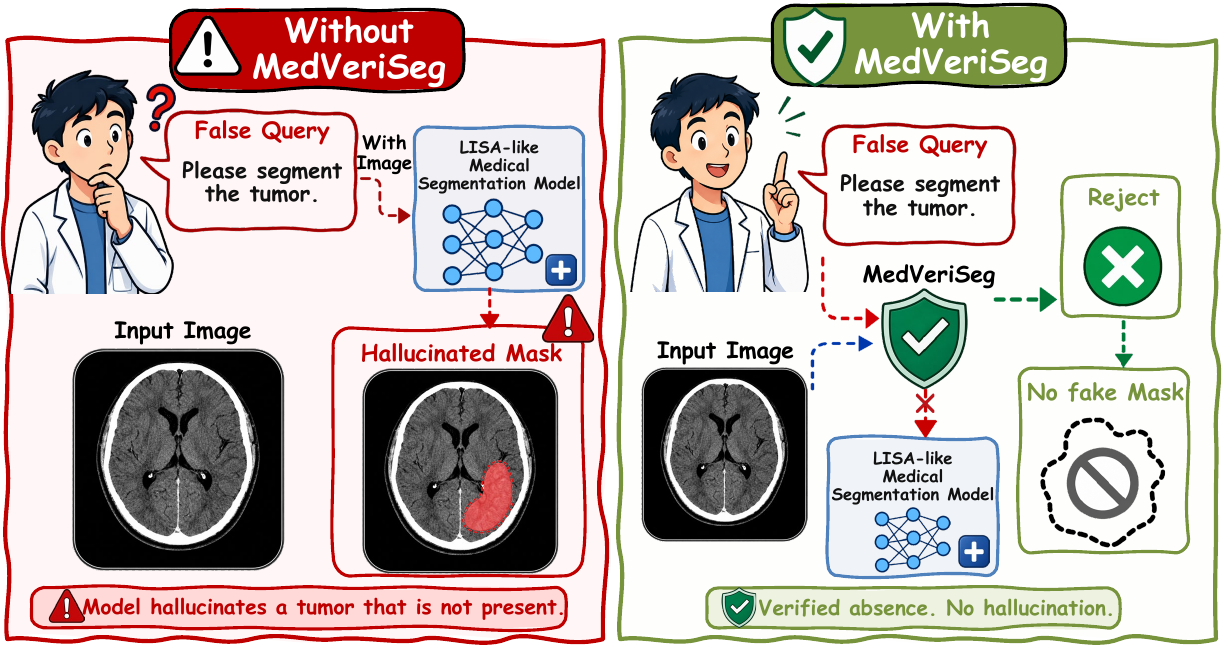}
    \vspace{-20pt}
    \caption{ 
    Example of hallucinated segmentation by LISA-like models under false queries. MedVeriSeg enables these models to effectively identify false queries and prevent hallucinated mask generation.
    } 
    \label{fig:teaser}
\end{figure}

\begin{figure*}[t]
    \centering
    \includegraphics[width=1\linewidth]{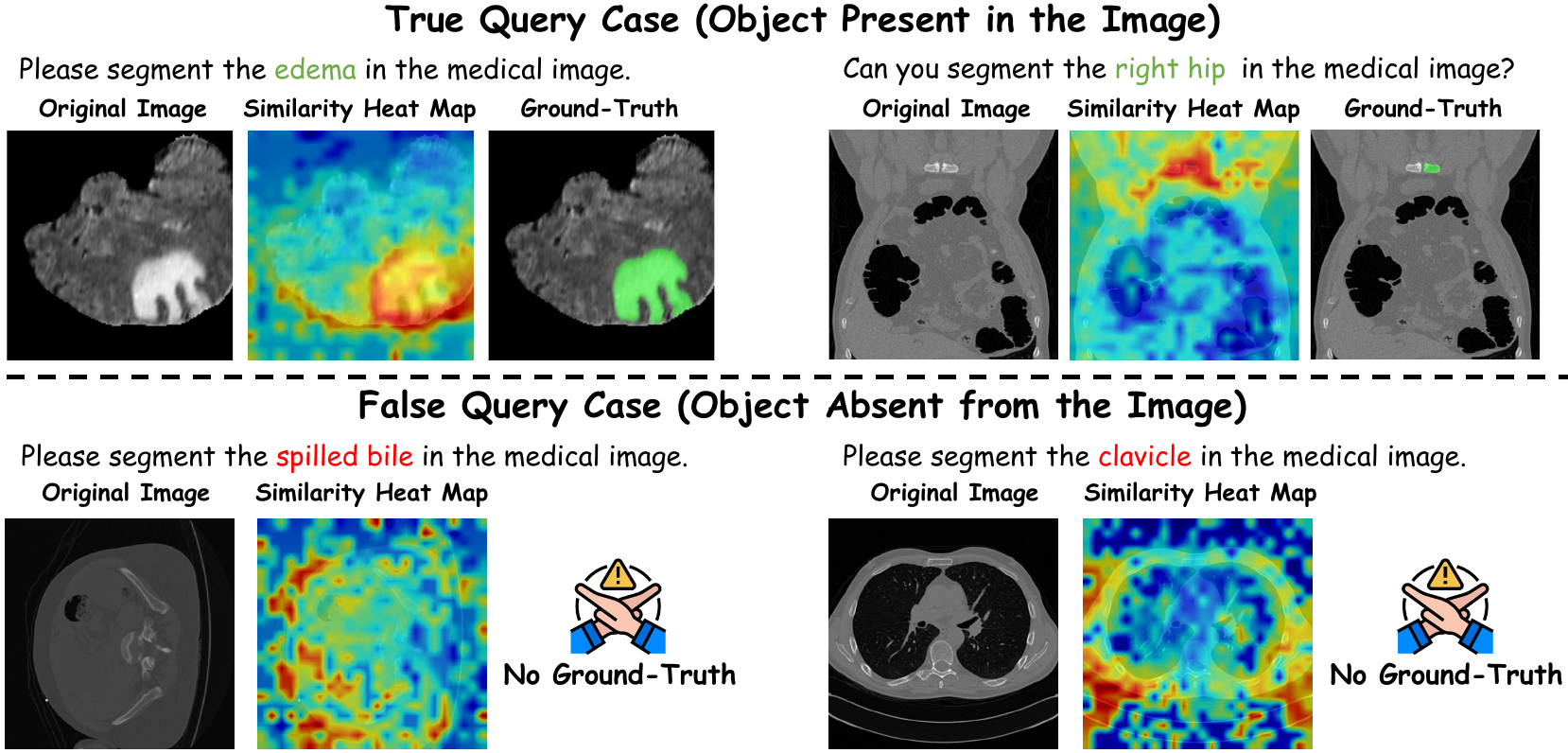}
    \vspace{-20pt}
    \caption{Distribution patterns of similarity heatmaps under true and false query cases. 
    We use MediSee \cite{tong2025medisee} as backbone for the visual analysis.
    The similarity matrix is computed between the hidden-state representation of the \texttt{[SEG]} token and the image features extracted from the final hidden layer of the MLLM.} 
    \label{fig:vis_1}
\end{figure*}

More recently, several MLLM-based medical segmentation models \cite{tong2025medisee,huang2025towards_MedPLIB,huang2025medsegR,yan2025medreasoner,tong2025mediround,nature-llm-seg} have further advanced model interactivity.
In addition to handling simple textual queries, these models can interpret and respond to more complex instructions, thereby substantially improving their practicality and efficiency in clinical and medical education scenarios.
From an architectural perspective, they commonly follow a LISA-like paradigm \cite{lisa}, in which an MLLM (\textit{e.g.}, LLaVA-Med \cite{llava-med} or LLaVA \cite{llava}) jointly reasons over the medical image and the input query, and then generates a special token (\textit{e.g.}, \texttt{[SEG]}) to activate a downstream segmentation decoder (\textit{e.g.}, MedSAM \cite{medsam_model} or SAM-Med2D \cite{sam-med-2d-model}) for mask prediction.
Although the resulting segmentation performance is highly competitive under such a well-crafted architecture design, these medical segmentation models almost invariably suffer from a common limitation: when confronted with false instructions (\textit{i.e.}, cases in which the target object specified in the instruction is actually absent from the medical image), they fail to recognize that the request should be refused and instead still produce the \texttt{[SEG]} token, thereby performing hallucinated segmentation, as illustrated in Figure~\ref{fig:teaser}.
This limitation can have serious consequences in medical image segmentation: in medical education, hallucinated outputs may distort students’ understanding of anatomical and pathological structures, while in clinical practice, they may contribute to misdiagnosis.
Addressing this issue is therefore essential for improving the practical reliability of these methods, while also reducing unnecessary downstream segmentation computations and saving computational resources.
Following prior general-domain work \cite{gsva_existing_work_general,liu2023gres_existing_work_general_2,glamm-cvpr2024,multi-round-seg}, our initial attempt is to adopt a retraining-based strategy, i.e., incorporating an appropriate number of negative samples into the training dataset to enable the model to learn such intelligent judgments.
However, this strategy suffers from two major drawbacks: \textit{(1)} the cost of data reconstruction and model retraining is prohibitively high, and \textit{(2)} the model may overfit to the newly constructed training data, resulting in limited generalizability.
We also explore the use of existing open-source general MLLMs \cite{bai2023qwen,deepseek-janus} as a prior module to assess whether the entities mentioned in the segmentation query are present in medical images.
However, the results remain unsatisfactory. 
This may be attributed to the limited ability of general MLLMs to perceive fine-grained visual cues in medical images, as well as the unreliability of direct qualitative judgments in such complex reasoning tasks.
Therefore, we raise the following question: \textit{Can we leverage the structural characteristics of LISA-like medical image segmentation models to develop a lightweight yet reliable verification framework that enables false-query rejection without additional training, thereby mitigating hallucinated segmentation?}

Existing studies on LISA-like methods have demonstrated that the information encoded by the \texttt{[SEG]} token directly affects the resulting segmentation masks \cite{tong2025medisee,lisa-READ}. 
Motivated by this observation, we further investigate whether the feature distribution of the \texttt{[SEG]} token exhibits distinguishable patterns under true and false medical image segmentation queries.
We draw on the visualization strategies adopted in READ \cite{lisa-READ} to analyze the feature distribution of the \texttt{[SEG]} token. 
Specifically, we compute the similarity matrix between the \texttt{[SEG]} token feature and the image features derived from the last hidden layer of the MLLM output, and then visualize the resulting similarity matrix as a heatmap. 
As shown in Figure \ref{fig:vis_1}, when presented with a valid query, the \texttt{[SEG]} token tends to attend to the true target region and its surrounding areas, producing a heatmap with clear concentration and well-defined intensity patterns.
In contrast, when the queried object is absent from the medical image, the heatmap typically appears irregular, diffuse, and unfocused.

\begin{figure*}[t]
    \centering
    \includegraphics[width=1\linewidth]{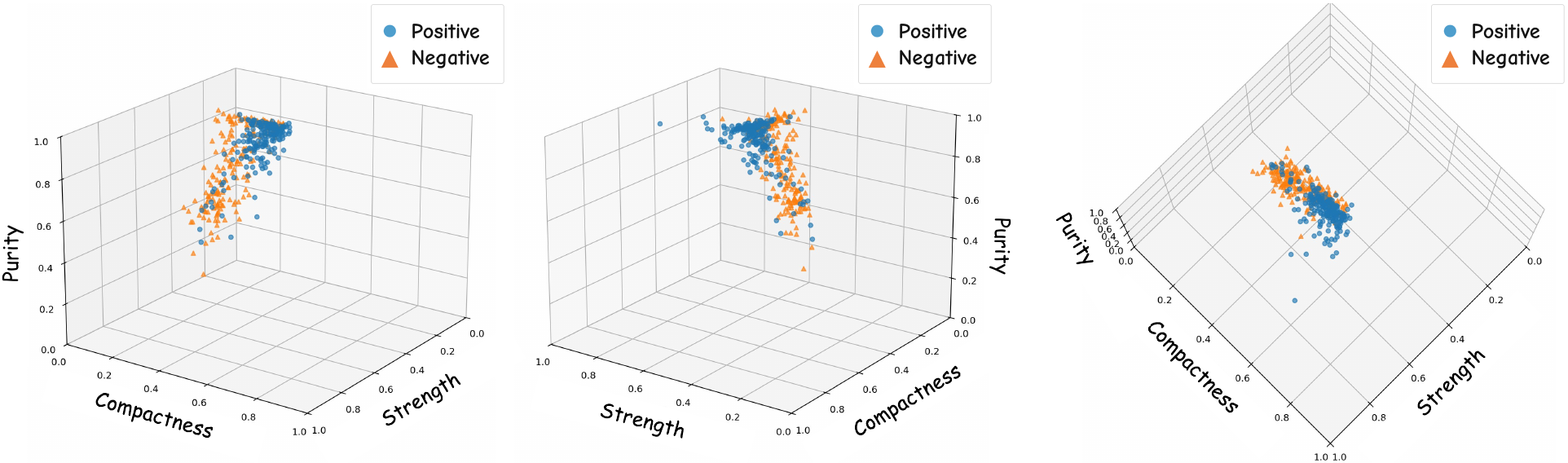}
    \vspace{-20pt}
    \caption{Distributions of 200 positive and 200 negative samples across the three SRQS dimensions: \textit{strength}, \textit{compactness}, and \textit{purity}. 
    The local overlap between positive and negative samples indicates limited separability under SRQS alone.} 
    \label{fig:sample_distribution}
\end{figure*}

Motivated by this finding, we first design a \textbf{Similarity Response Quality Scoring Module} (SRQS). This module quantitatively characterizes the distribution of the aforementioned similarity matrix from three perspectives, \textit{i.e.}, \textit{strength}, \textit{compactness}, and \textit{purity}, thereby determining whether the target object specified by the query is present in the medical image.
However, directly using SRQS as a decision criterion yields suboptimal performance, with approximately 60\% accuracy in a small-scale validation experiment. 
Figure \ref{fig:sample_distribution} further analyzes this behavior by showing the distributions of 200 positive and 200 negative samples across the three SRQS dimensions. 
Although most samples can be separated, the local overlap between the two groups suggests that these indicators alone are insufficient, likely due to the inherent image-specificity of query responses in medical images.
Based on this analysis, we further complement the quantitative evidence from SRQS with qualitative visual evidence from the raw image and the similarity heatmap, enabling a more comprehensive and robust multi-source verification.
Specifically, we propose \textbf{MedVeriSeg}, which no longer treats SRQS as the sole decision-maker, but instead incorporates its outputs into a \textbf{Lightweight Routed Multi-Agent Verification Module} (LMAV) as key quantitative evidence. 
LMAV leverages the strengths of MLLMs in modular qualitative analysis and multi-source information integration \cite{zhang2023multimodalcot,yang2023mm,suris2023vipergpt}. 
It collects evidence cards from different agents, including the SRQS Agent, the Raw Image Semantic Agent, and the Heatmap Distribution Agent, and integrates both quantitative and qualitative evidence to produce a comprehensive and rigorous assessment of the validity of the medical image segmentation query.


Considering the lack of datasets specifically designed to evaluate the ability of LISA-like medical segmentation models to distinguish between true and false segmentation queries, we construct \textbf{MedVeriSeg-Bench}, a benchmark for systematically assessing the effectiveness of our method and supporting future related research.
Extensive experiments demonstrate that, equipped with the proposed MedVeriSeg framework, LISA-like medical image segmentation models can effectively identify whether the target specified by a segmentation query is present in the medical image. 
This not only prevents users from receiving incorrect hallucinated segmentation results but also reduces unnecessary computational overhead in the downstream decoder.
Notably, MedVeriSeg requires no additional training and can be readily integrated into various LISA-like medical image segmentation models as a plug-in module, demonstrating its flexibility and practical applicability.
In summary, our  main contributions are as follows:
\begin{itemize}
    \item 
    We identify and formulate the hallucinated segmentation in LISA-like medical image segmentation models, where models may still produce masks for false queries whose target objects are absent from the image. 
    
    \item  We propose \textbf{MedVeriSeg}, an effective query verification framework for LISA-like medical image segmentation models. 
    MedVeriSeg exploits the characteristics of the \texttt{[SEG]} token and introduces a Similarity Response Quality Scoring Module together with a Lightweight Routed Multi-Agent Verification Module to determine whether a queried target is truly present in the image.
    
    \item We construct \textbf{MedVeriSeg-Bench} to evaluate query verification in medical image segmentation. Experiments demonstrate that, for existing LISA-like models, MedVeriSeg reduces hallucinated segmentation while largely preserving their original segmentation capability.
\end{itemize}

The remainder of this paper is organized as follows. 
Section \ref{sec:related} provides a comprehensive review of the related work. 
Sections \ref{sec:method} and \ref{sec:data_Pipeline} then present the proposed method and benchmark in detail. 
Extensive experiments are conducted in Section \ref{sec:exp} to comprehensively evaluate the performance of our method. 
Limitations are discussed in Section \ref{sec:limitation}.
Finally, Section \ref{sec:conclusion} concludes the paper.
We provide the code for our method to facilitate readers’ reference to implementation details that are not fully covered in the main paper.

\section{Related Work} \label{sec:related}
\subsection{MLLM-based Medical Image Segmentation}

Recent advances in multimodal large language models (MLLMs) \cite{gpt-5,perceptiongpt-cvpr2024,tong2025chat,csvt_mllm_1,csvt_mllm_2} have promoted a new paradigm of interactive medical image segmentation (MLLM-based), where textual instructions are used to specify target regions beyond conventional category or box-based prompts.
Early efforts, such as LLMSeg~\cite{nature-llm-seg}, incorporate clinical textual information into target volume contouring for radiation oncology, showing the potential of language-driven medical delineation.
More recently, several works further equip biomedical MLLMs with pixel-level perception ability.
MedPLIB~\cite{huang2025towards_MedPLIB} supports biomedical visual question answering, pixel-level prompting, and grounding within a unified MLLM framework.
MediSee~\cite{tong2025medisee}, MedSeg-R~\cite{huang2025medsegR}, and MedReasoner~\cite{yan2025medreasoner} extend medical image segmentation from explicit referring expressions to implicit or clinically motivated reasoning queries, enabling models to infer target regions from more complex textual instructions.
MediRound~\cite{tong2025mediround} further explores multi-round reasoning segmentation in medical images.
Despite these advances, most existing MLLM-based medical segmentation methods generally adopt a LISA-like paradigm~\cite{lisa,lisa++}, which relies on the \texttt{[SEG]} token to activate mask decoding.
However, such models are typically trained on substantial valid query-mask pairs and may therefore still emit \texttt{[SEG]} for false queries, leading to hallucinated segmentation when the queried target is absent.
In this work, we focus on this overlooked reliability issue and study how to enable LISA-like medical segmentation models to verify whether the queried target truly exists before producing a mask.

\subsection{Segmentation Query Identification}

In the general domain, several studies have explored segmentation under invalid or false-premise textual queries.
GRES~\cite{liu2023gres_existing_work_general_2} extends conventional referring expression segmentation to a generalized setting, where a query may refer to single, multiple, or no target objects.
Following this setting, GSVA~\cite{gsva_existing_work_general} introduces a \texttt{[REJ]} token into MLLM-based segmentation to explicitly reject null-target queries.
SESAME~\cite{see_say_segment} further studies false-premise reasoning segmentation, enabling models to determine whether the queried object exists before producing a mask.
In contrast, related exploration in medical image segmentation remains limited.
Existing medical-domain relevant methods~\cite{tang2025ltse,choi2026instruction}  are mainly designed for specific scenarios and typically rely on specific data construction and large-scale model training, without explicitly addressing the false-query limitation of LISA-like medical segmentation models as a general problem.
To address this gap, we propose a training-free query verification framework that can serve as a plug-in for different LISA-like medical segmentation models.

\section{Method} \label{sec:method}

\subsection{Revisiting LISA-Style Segmentation Structure}
\label{sec:revisiting}
\begin{figure*}[t]
    \centering
    \includegraphics[width=1\linewidth]{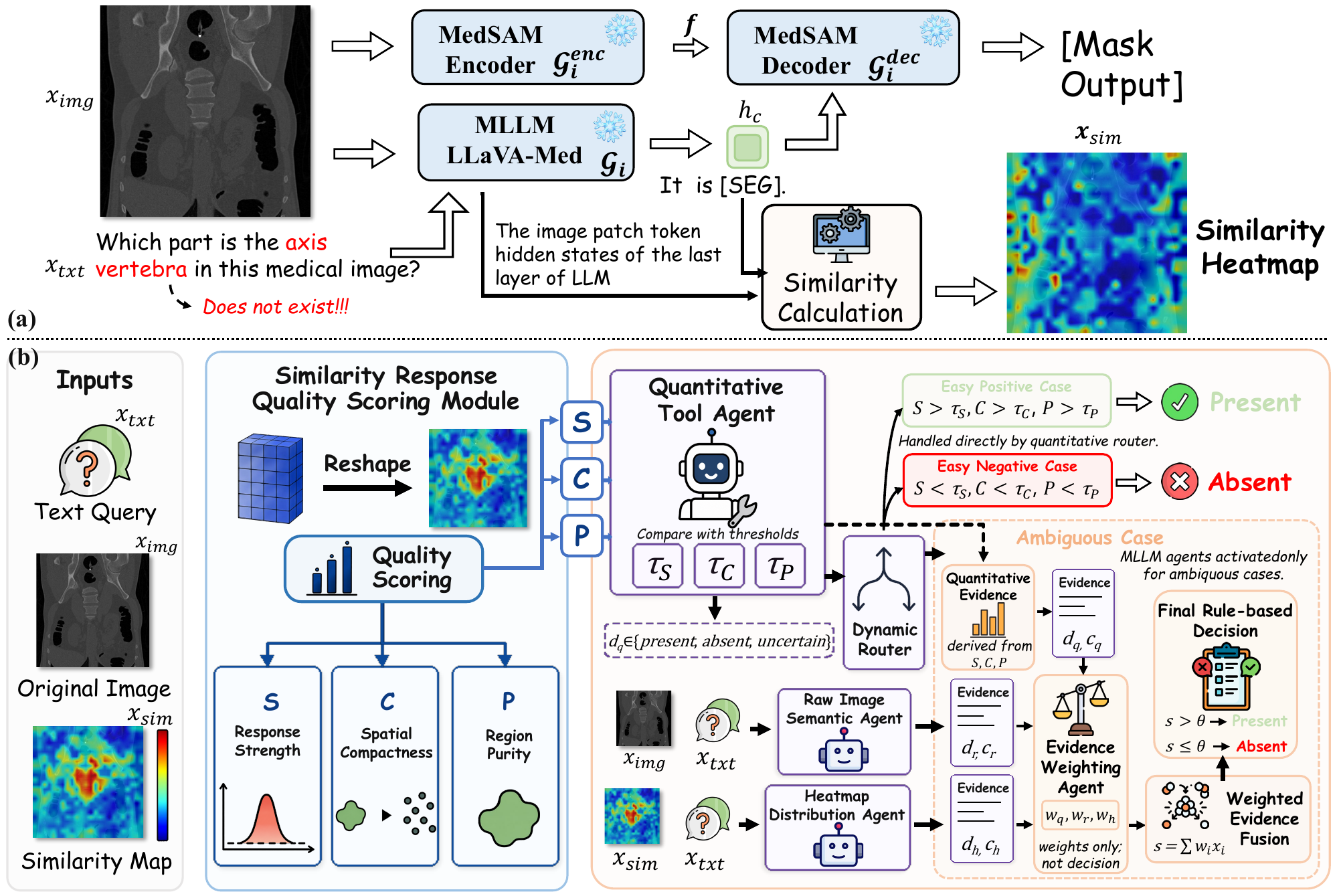}
    \vspace{-20pt}
    \caption{
    Revisiting the LISA-style segmentation structure (a), using MediSee~\cite{tong2025medisee} as an example, and overview of the proposed MedVeriSeg framework (b).
    MedVeriSeg consists of two main components: the Similarity Response Quality Scoring Module (SRQS) and the Lightweight Routed Multi-Agent Verification Module (LMAV).
       SRQS derives quantitative evidence from the similarity matrix, while LMAV integrates it with qualitative evidence from other agents for comprehensive analysis.
   } 
    \label{fig:method}
\end{figure*}

Figure~\ref{fig:method} (a) illustrates the main components of the existing  classical LISA-like medical image segmentation architecture, using MediSee \cite{tong2025medisee} as an example. The architecture comprises three core modules: MedSAM~\cite{medsam_model}, serving as the vision backbone $\mathcal{G}_i^{enc}$ and mask decoder $\mathcal{G}_i^{dec}$, and LLaVA-Med~\cite{llava-med}, serving as the MLLM $\mathcal{G}_i$.

Specifically, these methods first augment the original LLM vocabulary with a dedicated \texttt{[SEG]} token, emitted as a marker token when the model is prompted for medical image segmentation.
As illustrated in Figure~\ref{fig:method} (a), given a textual instruction \(x_{txt}\) and an input image \(x_{img}\), both inputs are fed into LLaVA-Med \(\mathcal{G}_i\), which then produces a textual response \(\hat{y}_{txt}\).
It can be formulated as:
\begin{align}
\begin{aligned}
    \hat{y}_{txt} = & \;  \mathcal{G}_i(x_{img}, x_{txt}).
\end{aligned}
\end{align}
The output \(\hat{y}_{txt}\) contains the \texttt{[SEG]} token whenever the model is intended to perform segmentation.
They then extract the last-layer LLM embedding associated with the \texttt{[SEG]} token \(h_{c}\).
Finally, the dense visual features \(f\), extracted from the input image \(x_{img}\) by the vision backbone \(\mathcal{G}_i^{enc}\), are fed into the decoder \(\mathcal{G}_i^{dec}\) together with \(h_{c}\) to produce the segmentation mask \(\hat{\mathbf{M}}\).
The process can be formulated as:
\begin{align}
\begin{aligned}
     f = \mathcal{G}_i^{enc}&\;(x_{img}), \quad \hat{\mathbf{M}} = \mathcal{G}_i^{dec}(f, h_{c}).
\end{aligned}
\end{align}

The above procedure constitutes the classical LISA-style segmentation pipeline for medical images.
Although this pipeline can achieve relatively high segmentation accuracy, it is inherently limited in that the MLLM \(\mathcal{G}_i\) tends to output the \texttt{[SEG]} token and subsequently perform segmentation regardless of whether the target object specified in the query \(x_{txt}\) is actually present in the medical image \(x_{img}\).
Throughout this process, the model often lacks the ability to assess the validity of the query, thereby producing hallucinated segmentation results.
In the following section, we will introduce the proposed MedVeriSeg to mitigate this issue.

\subsection{MedVeriSeg Framework}
\label{sec:MedVeriSeg}
As discussed in Sec.~\ref{sec:intro}, whether the target specified by the query \(x_{txt}\) is present in the medical image leads to clearly distinguishable distribution patterns in the heatmap of the similarity matrix between the hidden states of the \texttt{\texttt{[SEG]}} token and the visual feature embeddings in the MLLM output.
This visual observation suggests that the similarity matrix contains information cues for target-presence identification. 
However, limiting the analysis of the heatmap to qualitative assessment of its visual distribution lacks sufficient rigor and robustness.
Therefore, we first formalize this phenomenon by quantitatively characterizing the value distribution of the similarity matrix.
To obtain a more comprehensive analytical judgment, we then introduce the similarity matrix quantitative results as key quantitative evidence into a lightweight multi-agent collaborative verification module.

Overall, as shown in Figure~\ref{fig:method} (b), the MedVeriSeg framework consists of two main components. The first is the Similarity Response Quality Scoring Module (SRQS), which computes quantitative evidence from the similarity response. 
The second is the Lightweight Routed Multi-Agent Verification Module (LMAV), which integrates the quantitative evidence with multi-agent verification to obtain the final prediction.

\subsubsection{Similarity Response Quality Scoring Module}
We first design a scoring module based on three intuitive criteria: \emph{response strength} $S$, \emph{spatial compactness} $C$, and \emph{region purity} $P$. 
The main intuition is that a true target tends to induce the similarity map to exhibit not only strong responses in its vicinity, but also a compact and coherent high-response region.

Let the input similarity tensor be reshaped into a 2D response map \(M\):
\begin{equation}
M \in \mathbb{R}^{H \times W}.
\end{equation}
We first compute two robust statistics from all entries in \(M\): the median \(q_{50}\) and the upper quantile \(q_{95}\). 
They are used to normalize the response distribution and reduce sensitivity to scale variations.

\noindent \myparagraph{Response Strength.}
We measure whether the highest responses stand out clearly from the background. Specifically, we take the mean value of the top-\(k\) responses in \(M\):
\begin{equation}
m_{top} = \mathrm{mean}(\mathrm{TopK}(M)),
\end{equation}
where \(k=\max(1,\lfloor \rho HW \rfloor)\).
Then a normalized strength value is defined as:
\begin{equation}
r_s = \max\left(0,\frac{m_{top}-q_{50}}{q_{95}-q_{50}+\varepsilon}\right),
\end{equation}
where \(\varepsilon\) is a small constant for numerical stability.
To map \(r_s\) into \([0,1]\), we apply a sigmoid function:
\begin{equation}
s_1 = sigmoid({r_s}).
\end{equation}
If the top responses are significantly higher than the background level, \(s_1\) becomes large; otherwise it stays low.

\noindent \myparagraph{Spatial Compactness.}
A true target is expected to produce a spatially concentrated response pattern rather than a broadly scattered one. To evaluate this property, we first convert the raw response map \(M\) into a non-negative score map:
\begin{equation}
S = \max\left(0,\frac{M - q_{50}}{q_{95} - q_{50} + \varepsilon}\right).
\end{equation}
Then a small average smoothing kernel is applied to \(S\) to suppress isolated noise. Based on the resulting score map, we define an active region by thresholding:
\begin{equation}
\delta = \max\bigl(Q_{\alpha}(S_{>0}), \delta_{\min}\bigr),
A = \{(i,j)\mid S(i,j)\geq \delta\},
\end{equation}
where \(\delta_{\min}\) is a minimum activation threshold used to suppress weak scattered responses and \(Q_{\alpha}(\cdot)\) denotes the \(\alpha\)-quantile of positive scores. 
This step retains relatively strong responses while suppressing weak scattered activations.
We then compute the weighted center of the active region:
\begin{equation}
\mathbf{c} =
\frac{\sum_{(i,j)\in A} S(i,j)\mathbf{p}_{ij}}
{\sum_{(i,j)\in A} S(i,j) + \varepsilon},
\end{equation}
where \(\mathbf{p}_{ij}\) is the coordinate of location \((i,j)\).
The normalized weighted spread is defined as:
\begin{equation}
d =
\frac{\sum_{(i,j)\in A} S(i,j)\|\mathbf{p}_{ij}-\mathbf{c}\|_2}
{\left(\sum_{(i,j)\in A} S(i,j)\right)D + \varepsilon},
\end{equation}
where \(D\) is the diagonal length of the map for normalization. The compactness score is then defined as:
\begin{equation}
s_2 = \exp\left(-\frac{d}{\tau_c}\right),
\end{equation}
where \(\tau_c\) is a temperature parameter controlling the decay rate of the compactness score.
If the active responses are concentrated around a single center, the spread \(d\) is small and the compactness score \(s_2\) becomes high. In contrast, spatially dispersed responses lead to a lower \(s_2\).

\noindent \myparagraph{Region Purity.}
Using the score map \(S\) and active region \(A\) defined above, we further assess whether the active responses are dominated by a single coherent region. Specifically, let \(C_{\max}\) denote the dominant connected component in \(A\). The purity score is defined as:
\begin{equation}
s_3 =
\frac{\sum_{(i,j)\in C_{\max}} S(i,j)}
{\sum_{(i,j)\in A} S(i,j) + \varepsilon}.
\end{equation}
If most of the active energy is concentrated in one dominant connected region, \(s_3\) is close to 1. Otherwise, if the responses are fragmented into multiple disconnected regions, \(s_3\) becomes smaller.
For simplicity, we denote the three scores as $S=s_1$, $C=s_2$, and $P=s_3$.
\textit{Further hyperparameter configurations and implementation details of this module can be found in the provided code.}


\subsubsection{Lightweight Routed Multi-Agent Verification}
Directly applying the Similarity Response Quality Scoring Module to the task yields suboptimal performance. Specifically, on a manually curated small-scale test set containing 101 samples (positive and negative in total), the module correctly identifies only 57 samples.
We observe that, although the visualized heatmap distributions of the similarity matrix indeed exhibit different characteristics under true and false queries, the response pattern for each medical image is highly image-specific due to the complex fusion between its visual features and the corresponding textual prompt. 
As a result, threshold-based discrimination using the three defined metrics is insufficient to handle such diverse and exceptional cases.

Therefore, we introduce the quantitative results of SRQS as evidence into a multi-agent collaborative reasoning module, enabling a comprehensive analysis that integrates both quantitative and qualitative evidence.
Specifically, as shown in Figure~\ref{fig:method} (b), given a query $x_{txt}$, an original image $x_{img}$, a heatmap $x_{sim}$, and three quantitative indicators, \textit{i.e.}, response strength $S$, compactness $C$, and purity $P$, the Quantitative Tool Agent first produces a quantitative decision as:
\begin{align}
\begin{aligned}
d_q =
\begin{cases}
\text{present}, & S>\tau_S,\ C>\tau_C,\ P>\tau_P, \\
\text{absent}, & S<\tau_S,\ C<\tau_C,\ P<\tau_P, \\
\text{uncertain}, & \text{otherwise}.
\end{cases}
\end{aligned}
\end{align}

A dynamic router is then used to reduce unnecessary agent calls. 
Specifically, if all three indicators are above their thresholds, the system directly outputs \textit{present}. If all three indicators are below their thresholds, the system directly outputs \textit{absent}. 
Otherwise, the case is regarded as ambiguous and routed to visual verification agents.

For ambiguous cases, two visual agents are activated. 
Specifically, the raw image semantic agent checks whether the original medical image $x_{img}$ provides visual evidence for the queried target specified by $x_{txt}$. 
The heatmap distribution agent evaluates whether the heatmap $x_{sim}$ response is spatially focused and consistent with a plausible target region. 
Each agent outputs an evidence card containing a decision $d_i \in \{\text{present}, \text{absent}, \text{uncertain}\}$ and a confidence score $c_i \in [0,1]$.
For \(i\), it is the evidence sources:
\begin{align}
\begin{aligned}
i \in \{q, r, h\},
\end{aligned}
\end{align}
where $q$, $r$, and $h$ denote the quantitative evidence, raw image evidence, and heatmap evidence, respectively. 
For the Quantitative Tool Agent, the confidence score $c_q$ is defined based on the distance between the quantitative similarity metrics and their corresponding decision thresholds. 
The confidence score is first computed as the mean absolute margin:
\begin{align}
\begin{aligned}
\tilde{c}_q
&= \frac{1}{3}
\left(
|S-\tau_S|
+
|C-\tau_C|
+
|P-\tau_P|
\right).
\end{aligned}
\end{align}
Then it is clipped to the range $[0,1]$:
\begin{align}
\begin{aligned}
c_q &= \operatorname{clip}_{[0,1]}(\tilde{c}_q).
\end{aligned}
\end{align}
Finally, the confidence score is determined according to the quantitative decision $d_q$, specifically:
\begin{align}
\begin{aligned}
c_q &=
\begin{cases}
\min\left(
\operatorname{clip}_{[0,1]}(\tilde{c}_q),
\gamma
\right),
& d_q=\text{uncertain}, \\
\operatorname{clip}_{[0,1]}(\tilde{c}_q),
& d_q\in\{\text{present},\text{absent}\}.
\end{cases}
\end{aligned}
\end{align}
This design makes $c_q$ reflect how far the quantitative evidence is from the decision boundary, rather than directly representing the probability of target presence. A larger $c_q$ indicates that the quantitative metrics are farther from their thresholds and thus provide stronger evidence. The upper bound $\gamma$ for uncertain cases prevents internally inconsistent quantitative evidence from dominating the final evidence fusion process.

To effectively integrate and balance the contributions from different evidence cards to the final decision, we employ an evidence weighting agent that assigns reliability weights to different evidence sources.
It is worth noting that the weighting agent only predicts the weights and does not directly make the final decision. 
The weights satisfy:
\begin{align}
\begin{aligned}
\sum_i w_i = 1, \quad w_i \geq 0.
\end{aligned}
\end{align}
Then each evidence decision is converted into an existence value $x_i$:
\begin{align}
\begin{aligned}
x_i =
\begin{cases}
0.5 + 0.5c_i, & d_i=\text{present}, \\
0.5, & d_i=\text{uncertain}, \\
0.5 - 0.5c_i, & d_i=\text{absent}.
\end{cases}
\end{aligned}
\end{align}
The final existence score is computed by weighted fusion:
\begin{equation}
s = \sum_i w_i x_i.
\end{equation}
The final decision is obtained by comparing $s$ with a threshold $\theta$:
\begin{align}
\begin{aligned}
y =
\begin{cases}
\text{present}, & s > \theta, \\
\text{absent}, & s \leq \theta.
\end{cases}
\end{aligned}
\end{align}
In this lightweight multi-agent design, the MLLM interface is used only for visual evidence analysis and evidence weighting. 
Notably, the final decision is not directly generated by the MLLM agent; instead, it is computed through an explicit score-based rule. This design makes the verification process more controllable and transparent.

\begin{figure*}[t]
    \centering
    \includegraphics[width=1\linewidth]{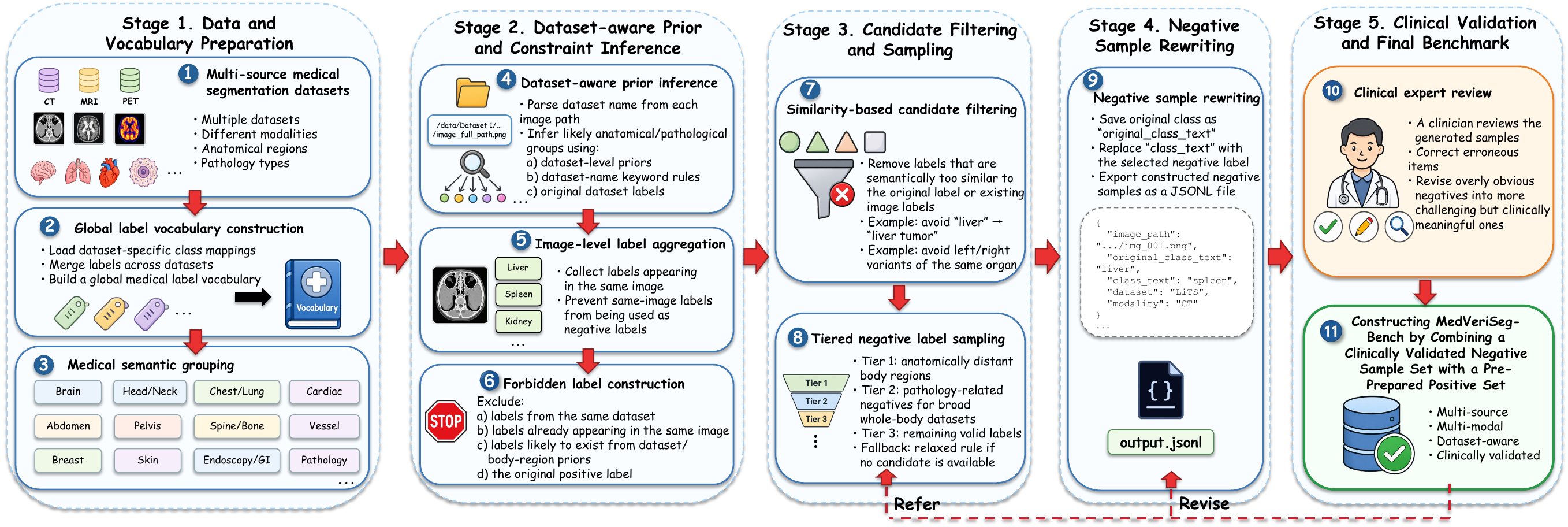}
    \vspace{-20pt}
    \caption{
    Clinician-in-the-Loop Negative Sample Construction for MedVeriSeg-Bench, which consists of both positive and negative samples. The negative samples are constructed using the above automatic pipeline, followed by clinician review to correct erroneous samples and refine overly simple samples.
    } 
    \label{fig:dataset_pipeline}
\end{figure*}

\begin{figure}[t]
    \centering
    \includegraphics[width=1\linewidth]{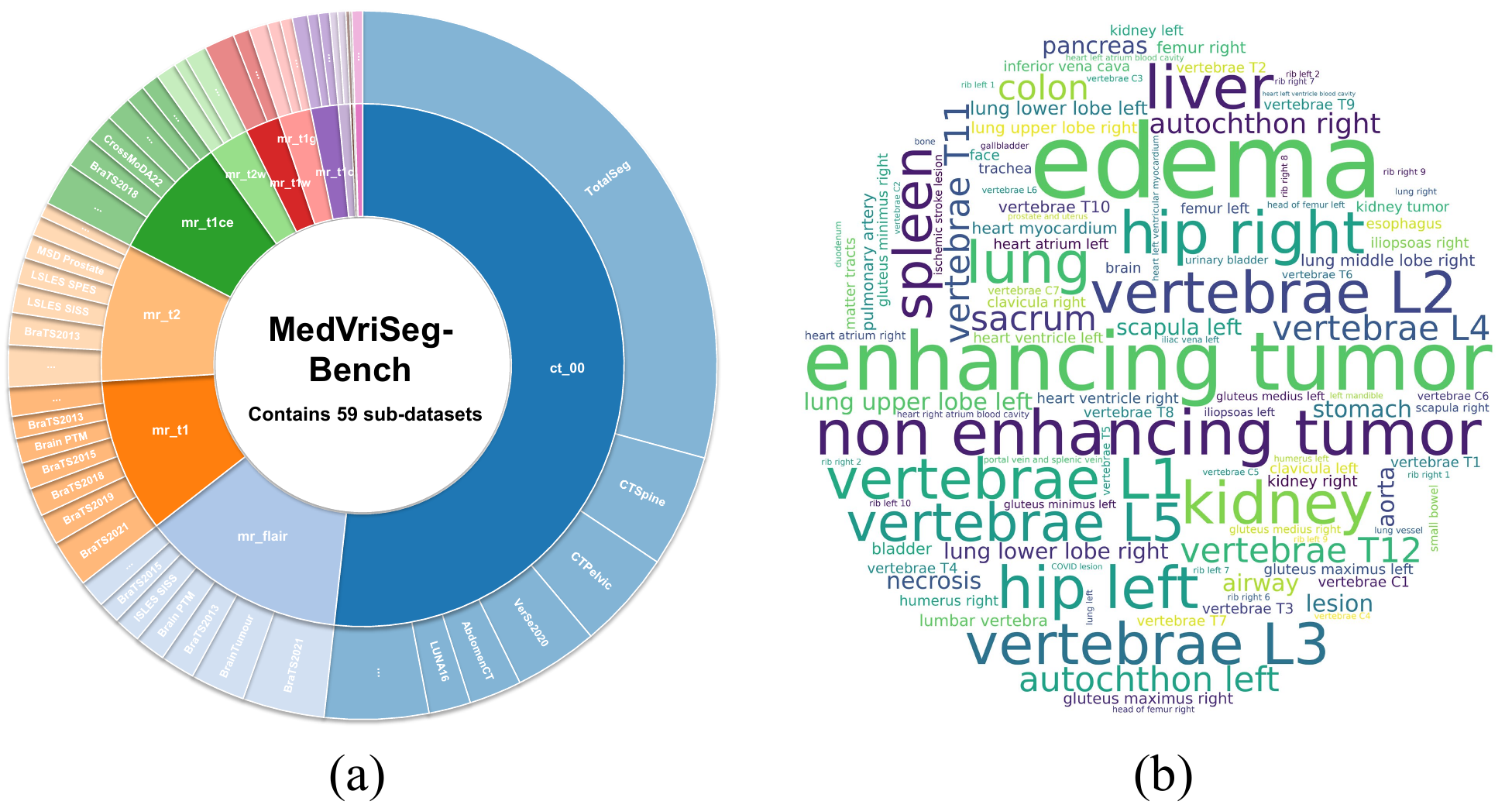}
    \vspace{-1.8em}
    \caption{Analysis of the data structure in MedVeriSeg-Bench dataset. (a) Hierarchical distribution of imaging modalities and source datasets; (b) Word cloud of MedVeriSeg-Bench. \textit{Please zoom in for improved clarity}.}
    \vspace{-1.2em}
    \label{fig:word_cload}
\end{figure}

\section{MedVeriSeg-Bench}\label{sec:data_Pipeline}
Given the scarcity of datasets designed to evaluate whether medical segmentation models or methods can distinguish between true and false textual queries, we construct MedVeriSeg-Bench to enable a more systematic and quantitative assessment of query authenticity verification capability.

\subsection{Data Construction Pipeline}
\noindent \myparagraph{Positive Sample Construction.}
We randomly sample a fixed number of positive examples from 59 public medical image segmentation datasets, with the sampling probability proportional to both dataset size and the real-world prevalence of different sample types. The construction of positive examples is straightforward: for each selected image-mask pair, the original class label of the mask is retained as its positive label.
During testing, the evaluated method is required to determine whether the textual segmentation instruction corresponding to each label truly refers to an object present in the associated medical image.

\noindent \myparagraph{Clinician-in-the-loop Negative Sample Construction.}
As illustrated in Figure \ref{fig:dataset_pipeline}, we design a clinician-in-the-loop pipeline to construct meaningful negative samples.
In \textit{Stage 1}, we first collect samples from the same pool of multi-source datasets used for positive sample construction, covering diverse imaging modalities, anatomical regions, and pathology types.
We then load dataset-specific class mappings and merge them into a unified global medical label vocabulary. 
To support anatomically aware negative sampling, all labels are further assigned to medical semantic groups (\textit{e.g.},  brain and head/neck).

Next, in \textit{Stage 2}, we perform dataset-aware prior and constraint inference. 
Specifically, we parse the dataset name from each sample and infer the anatomical or pathological groups that may appear in the image based on dataset-level priors, dataset-name keyword rules, and the labels originally contained in the dataset. 
Meanwhile, we aggregate image-level labels to prevent labels already present in the same image from being used as negative labels. 
Based on these constraints, we construct a forbidden label set by excluding labels from the same dataset, labels appearing in the same image, labels likely to exist according to dataset or body-region priors, and the original positive label.

Subsequently, in \textit{Stage 3}, we filter candidate negative labels by removing labels that are semantically too similar to the original label or existing image labels, such as fine-grained variants of the same organ or left/right counterparts. 
We then adopt a tiered sampling strategy: labels from anatomically distant body regions are preferred first, pathology-related labels are used for broad whole-body datasets, and remaining valid labels are sampled as fallback candidates. 
In \textit{Stage 4}, we rewrite each sample by preserving the original class and replacing the new class label with the selected negative label, and exporting the constructed samples into a JSONL file.

Considering that a pipeline without human involvement may produce erroneous samples, particularly overly simplistic ones, in \textit{Stage 5}, we employ three clinical experts to review the generated negative samples, with the goals of \textbf{correcting erroneous items} and \textbf{revising overly obvious negatives} into more challenging yet meaningful samples that can simulate plausible false queries arising in medical education or clinical practice.
By combining the negative sample set with a pre-prepared positive sample set, we obtain the final MedVeriSeg-Bench, a multi-source, multi-modal, and clinically validated benchmark for evaluating methods’ ability to infer the existence of true and false entities specified in queries for medical image segmentation.

\begin{table*}[t]
\footnotesize
\centering
\caption{
Comparison of true--false query verification performance between MedVeriSeg and baseline methods across major medical imaging modalities in MedVeriSeg-Bench. 
For MLLM-based methods, \textit{Image} and \textit{Heatmap} indicate the visual information used for verification, and \textit{Simple Query} and \textit{CoT Query} denote different levels of textual guidance. 
This table reports the overall verification performance across positive and negative samples.
The heatmap is produced by MediSee \cite{tong2025medisee}.
}
\label{table:main_table_modality}
\resizebox{\textwidth}{!}{%
\begin{tabular}{ l | c c | c c | c c | c c | c c }
\toprule[1.2pt]
\multirow{3}*{\textbf{Methods}} 
& \multicolumn{8}{c|}{\textbf{Modalities}} 
& \multicolumn{2}{c}{\textbf{All Set}} \\

\specialrule{0em}{0pt}{1pt}
\cline{2-11}
\specialrule{0em}{0pt}{1pt}

~ 
& \multicolumn{2}{c|}{\textbf{ct}} 
& \multicolumn{2}{c|}{\textbf{mr\_flair}} 
& \multicolumn{2}{c|}{\textbf{mr\_t1}} 
& \multicolumn{2}{c|}{\textbf{mr\_t1ce}} 
& \multicolumn{2}{c}{\textbf{overall}} \\

\specialrule{0em}{0pt}{1pt}
\cline{2-11}
\specialrule{0em}{0pt}{1pt}

~ 
& \textbf{Acc} & \textbf{F1}
& \textbf{Acc} & \textbf{F1}
& \textbf{Acc} & \textbf{F1}
& \textbf{Acc} & \textbf{F1}
& \textbf{Acc} & \textbf{F1} \\

\specialrule{0em}{0pt}{1pt}
\hline
\hline
\specialrule{0em}{0pt}{1pt}

\rowcolor{green!20}
\multicolumn{11}{c}{\textit{\textbf{Non-MLLM Verification Baselines}}} \\

Random Verification \cite{fawcett2006introduction} 
&50.29  &44.34  
&50.52  &55.78  
&50.26  &52.01  
&48.91  &57.14  
&50.37  &49.91  \\

Public-User Verification 
&43.61  &36.32  
&35.51  &40.19  
&43.48  &43.87  
&44.54  &49.40  
&42.05  &40.05  \\

Specialist-User Verification 
&90.05  &87.80  
&87.21  &89.44  
&82.96  &84.69  
&85.15  &87.55  
&88.23  &88.00  \\

\specialrule{0em}{0pt}{1pt}
\hline
\specialrule{0em}{0pt}{1pt}

\rowcolor{blue!20}
\multicolumn{11}{c}{\textit{\textbf{MLLM-based Methods}}} \\
LLaVA-Med \cite{llava-med} + Heatmap + CoT Query 
&61.10  &18.00  
&48.30  &38.32  
&52.35  &39.38  
&44.76  &28.73  
&55.65  &29.13  \\

DeepSeek-Janus-Pro \cite{deepseek-janus} + Image + CoT Query 
&68.10  &43.39  
&48.43  &32.71  
&44.52  &6.45  
&41.05  &16.15  
&57.37  &30.86  \\

LLaVA-Med \cite{llava-med} + Image + CoT Query 
&69.77  &48.41  
&49.09  &34.34  
&44.52  &9.12  
&42.79  &21.08  
&58.59  &34.78  \\

DeepSeek-Janus-Pro \cite{deepseek-janus} + Image + Simple Query 
&70.25  &51.89  
&47.91  &31.33  
&45.04  & 7.60 
&41.48  &15.72  
&58.60  &35.70  \\

LLaVA-Med \cite{llava-med} + Image + Simple Query 
&70.19  &56.20  
&51.04  &38.83  
&46.78  &15.00  
&45.63  &29.06  
&60.18  &43.18  \\

Qwen-3-VL \cite{bai2023qwen} + Heatmap + CoT Query 
&61.10  &56.09  
&67.62  &77.66  
&61.22  &72.37  
&61.35  &72.98  
&62.17  &66.90  \\

DeepSeek-Janus-Pro \cite{deepseek-janus} + Heatmap + CoT Query 
&61.59  &44.20  
&72.19  &79.18  
&70.09  &76.57  
&66.81  &75.16  
&65.43  &64.75  \\

Qwen-3-VL \cite{bai2023qwen} + Image + Simple Query 
&62.55  &52.50  
&70.37  &71.87  
&61.91  &54.85  
&71.62  &74.21  
&65.48  &60.59  \\

Qwen-3-VL \cite{bai2023qwen} + Image + CoT Query 
&71.53  &60.49  
&64.62  &62.83  
&55.83  &40.38  
&62.66  &60.69  
&67.12  &58.49  \\




\specialrule{0em}{0pt}{1pt}
\hline
\specialrule{0em}{0pt}{1pt}

\rowcolor{red!20}
\multicolumn{11}{c}{\textit{\textbf{Our Methods}}} \\

MedVeriSeg (kimi-k2.6 \cite{team2025kimi})
&79.36  &75.65  
&83.29  &87.83  
&82.09  &86.26  
&81.88  &86.33  
&81.37  &82.36  \\

\rowcolor{aliceblue!60} MedVeriSeg (Qwen-3-VL \cite{bai2023qwen})
&80.91  &77.45  
&88.12  &91.31  
&85.57  &88.23  
&87.77  &90.88  
&83.93  &84.84  \\

\bottomrule[1.2pt]
\end{tabular}
}
\end{table*}

\begin{table}[]
  \centering
  \caption{Implementation details of MedVeriSeg (based on MediSee).}
  \resizebox{0.8\linewidth}{!}{%
  \begin{tabular}{l|c}
    \toprule
    Setting & Value \\
    \midrule
    Agent MLLM Backbone       & Qwen-3-VL \\
    Python Version            & 3.10.6 \\
    CUDA Version              & 11.7 \\
    GPU Type                  & NVIDIA RTX A6000 \\
    GPU Memory Usage          & about 19 GB \\
    Token Cost per Sample    & about 1600 tokens \\
    Inference Time per Sample & about 4 s \\
    \bottomrule
  \end{tabular}%
  }
  \label{tab:implementation_details}
\end{table}

\subsection{Data Statistics}
MedVeriSeg-Bench contains 6,022 query-image samples, consisting of 3,014 positive samples and 3,008 clinically validated negative samples.
The benchmark covers 151 medical entity categories and 24 imaging modalities, spanning a wide range of anatomical structures, lesions, and tissues.
To ensure a unified standard, we follow the modality classification scheme in SA-Med2D-20M \cite{sam_med_2d_20m}.
All samples are collected from 59 public medical segmentation datasets, providing broad coverage across heterogeneous data sources and different scenarios.

We further provide statistical analyses of MedVeriSeg-Bench to characterize its data distribution.
As shown in Figure \ref{fig:word_cload} (a), the sunburst chart illustrates the hierarchical distribution of the benchmark, where the inner ring denotes imaging modalities and the outer ring denotes the corresponding source datasets.
This visualization demonstrates the multi-source and multi-modal nature of MedVeriSeg-Bench. Moreover, both the modality distribution and the source-dataset distribution in MedVeriSeg-Bench exhibit a long-tailed pattern, reflecting the distributional characteristics of real-world medical imaging data.
In addition, the word cloud in Figure \ref{fig:word_cload} (b) presents the distribution of medical entity categories, highlighting the diversity of medical entities included in the benchmark.

\begin{table*}[t]
\centering
\small
\caption{
Category-wise comparison of true--false query verification performance between MedVeriSeg and baseline methods on MedVeriSeg-Bench. 
Accuracy (Acc.) is used as the evaluation metric. 
Each triplet denotes the accuracy on positive samples, negative samples, and all samples, respectively. 
This table reports results on a subset of 16 categories from the dataset.
For the compared baselines, the visual information is the image, and the query type is simple.
}
\label{table:main_table_class}
\setlength{\tabcolsep}{5pt}
\renewcommand{\arraystretch}{1.12}
\resizebox{\textwidth}{!}{
\begin{tabular}{l|ccccc}

\toprule
\multirow{3}{*}{\textbf{Category}} 
& \multicolumn{5}{c}{\textbf{Methods}} \\

& \textbf{DeepSeek-Janus-Pro} 
& \textbf{LLaVA-Med} 
& \textbf{Qwen-3-VL} 
& \textbf{MedVeriSeg (kimi-k2.6)} 
& \textbf{MedVeriSeg (Qwen-3-VL)} \\

& \cite{deepseek-janus}
& \cite{llava-med}
& \cite{bai2023qwen}
& Ours
& Ours \\

\specialrule{0em}{0pt}{1pt}
\hline
\hline
\specialrule{0em}{0pt}{1pt}

edema & (18.80, 99.64, 51.42) & (30.53, 97.47, 57.54) & (69.84, 91.88, 78.73) & (93.28, 69.13, 83.54) & (98.29, 73.29, 88.20) \\
enhancing tumor & (1.67, 100.00, 33.80) & (4.17, 97.00, 34.50) & (43.54, 91.85, 59.33) & (96.46, 69.53, 87.66) & (99.17, 71.24, 90.04) \\
non enhancing tumor & (1.07, 98.50, 41.65) & (3.48, 96.63, 42.28) & (31.55, 86.89, 54.60) & (89.04, 69.66, 80.97) & (91.98, 73.41, 84.24) \\
liver & (44.78, 91.45, 77.17) & (52.24, 89.47, 78.08) & (59.70, 79.61, 73.52) & (82.09, 85.53, 84.47) & (88.06, 79.61, 82.19) \\
hip right & (0.00, 97.22, 55.12) & (18.18, 94.44, 61.42) & (50.91, 86.11, 70.87) & (92.73, 84.72, 88.19) & (89.09, 90.28, 89.76) \\
hip left & (0.00, 91.67, 50.46) & (2.04, 85.00, 47.71) & (55.10, 81.67, 69.72) & (89.80, 81.67, 85.32) & (91.84, 85.00, 88.07) \\
kidney & (64.15, 87.76, 75.49) & (79.25, 87.76, 83.33) & (67.92, 89.80, 78.43) & (90.57, 85.71, 88.24) & (92.45, 79.59, 86.27) \\
lung & (51.85, 91.67, 79.31) & (51.85, 88.33, 77.01) & (51.85, 85.00, 74.71) & (70.37, 90.00, 83.91) & (70.37, 91.67, 85.06) \\
vertebrae L2 & (63.16, 88.89, 71.43) & (66.67, 77.78, 70.24) & (57.89, 70.37, 61.90) & (92.98, 96.30, 94.05) & (94.74, 88.89, 92.86) \\
spleen & (52.94, 89.80, 80.30) & (58.82, 87.76, 80.30) & (58.82, 77.55, 72.73) & (88.24, 79.59, 81.82) & (94.12, 81.63, 84.85) \\
autochthon left & (0.00, 93.18, 63.08) & (0.00, 88.64, 60.00) &(0.00, 59.09, 40.00) & (42.86, 72.73, 63.08) & (42.86, 79.55, 67.69) \\
vertebrae T12 & (54.29, 89.66, 70.31) & (65.71, 82.76, 73.44) & (37.14, 75.86, 54.69) & (97.14, 75.86, 87.50) & (94.29, 79.31, 87.50) \\
colon & (18.52, 91.43, 59.68) & (33.33, 85.71, 62.90) & (70.37, 68.57, 69.35) & (74.07, 80.00, 77.42) & (62.96, 74.29, 69.35) \\
vertebrae L4 & (69.05, 72.22, 70.00) & (76.19, 72.22, 75.00) & (73.81, 66.67, 71.67) & (95.24, 83.33, 91.67) & (100.00, 72.22, 91.67) \\
sacrum & (75.00, 94.29, 86.44) & (83.33, 91.43, 88.14) & (100.00, 80.00, 88.14) & (83.33, 91.43, 88.14) & (83.33, 91.43, 88.14) \\
vertebrae T11 & (44.83, 96.43, 70.18) & (48.28, 92.86, 70.18) & (41.38, 85.71, 63.16) & (89.66, 78.57, 84.21) & (93.10, 82.14, 87.72) \\

\specialrule{0em}{0pt}{1pt}
\hline
\hline
\specialrule{0em}{0pt}{1pt}

\textbf{Overall} 
& 58.60 & 60.18 & 65.48 & 81.37 & 83.93 \\
\bottomrule
\end{tabular}
}
\end{table*}

\section{Experiments}\label{sec:exp}
\subsection{Experimental Setup}
\label{sec:Experimental Setup dataset}


\subsubsection{Implementation Details}
Our method does not require any additional fine-tuning or parameter optimization.
In our main experiments, including all visualizations, we use MediSee, a representative work on medical MLLM-driven segmentation, as the backbone of the MedVeriSeg framework to compute the similarity matrix between the hidden states of \texttt{[SEG]} token and the image features from the LLM output layer.
For all experiments, the Lightweight Routed Multi-Agent Verification Module in MedVeriSeg is driven by Qwen-3-VL by default, using the API model \texttt{qwen3-vl-30b-a3b-instruct}.
All experiments are conducted on the proposed MedVeriSeg-Bench by default.
The implementation details of our framework are summarized in Table \ref{tab:implementation_details}.
Regarding the hyperparameters, in the multi-agent module, we set the thresholds for the three quantitative metrics as $\tau_S = 0.475$, $\tau_C = 0.4$, and $\tau_P = 0.7$; and the final decision threshold $\theta = 0.5$.
The upper bound $\gamma$ is set to \(0.65\).
\textit{SRQS hyperparameters and agent prompts in LMAV can be found in the provided code.}

\begin{table}[t]
\centering
\setlength{\belowcaptionskip}{4pt}
\caption{
Ablation study on the choice of agent MLLM backbones in the multi-agent module.
}
\label{table:ab_agent_mllm}
\small
\setlength{\tabcolsep}{3pt}
\resizebox{\columnwidth}{!}{
\begin{tabular}{l|cccc|c}
\toprule
\textbf{Method} & \textbf{Accuracy} & \textbf{F1} & \textbf{Precision} & \textbf{Recall} & \textbf{Average} \\
\midrule
\rowcolor{green!20}
\multicolumn{6}{c}{\textit{\textbf{Closed-source MLLMs}}} \\
GPT-4o~\cite{gpt-4}               &78.60 &79.52 &76.30  &83.01  & 79.36  \\
Gemini-2.5-Pro~\cite{team2023gemini}   &84.71 &85.49 &81.38  &90.05  &85.41  \\
\midrule
\rowcolor{blue!20}
\multicolumn{6}{c}{\textit{\textbf{Open-source MLLMs}}} \\

kimi-k2.6 \cite{team2025kimi}     &81.37 &82.36 &78.26  &86.93  &82.23  \\
\rowcolor{aliceblue!60}
Qwen-3-VL \cite{bai2023qwen}    &83.93 &84.84 &80.30  &89.95  &84.76  \\
\bottomrule
\end{tabular}
}
\end{table}

\subsection{True--False Query Verification Results}

The results of true–false query verification on MedVeriSeg-Bench are summarized in Tables \ref{table:main_table_modality} and \ref{table:main_table_class}. Specifically, Table \ref{table:main_table_modality} reports the performance of different methods across major modalities, whereas Table \ref{table:main_table_class} presents a category-wise analysis of their performance.
Since few prior studies have investigated the verification of hallucinated queries in LISA-like medical segmentation models, we first incorporate Random Verification and Human Verification to provide preliminary insights. 
Specifically, Random Verification denotes a fair random judgment of the veracity of each sample. 
Human Verification consists of Public-User Verification and Specialist-User Verification, corresponding respectively to evaluations conducted by general students without medical expertise and by medical students with domain knowledge to assess the veracity of the test samples.
As shown in Table \ref{table:main_table_modality}, Random Verification exhibits an accuracy trend close to 50\%. 
In contrast, Public-User Verification shows a clear performance drop due to the lack of medical expertise, which indirectly demonstrates the practical value (in medical education) of our task. 
Specialist-User Verification achieves an accuracy of 88\%, with stable F1 and Acc scores, indicating balanced performance on both positive and negative samples. 
However, such strong performance can be largely attributed to the evaluators’ extensive medical knowledge and requires substantial human effort.
In contrast, although the proposed MedVeriSeg does not surpass Specialist-User Verification, the gap is only approximately 5\%. 
Moreover, it substantially outperforms both Random Verification and Public-User Verification. 
Notably, our method is fully automated, requires no large-scale training and remains highly lightweight.

We further construct an MLLM-based paradigm for a more comprehensive comparison, leveraging the strong vision-language understanding capabilities of existing popular MLLMs \cite{bai2023qwen,deepseek-janus,llava-med} and employing textual guidance at different levels to support analysis and judgment based on different visual evidence.
Specifically, we sequentially provide the MLLMs with the original image and the similarity matrix heatmap as visual evidence, using either a simple query or a chain-of-thought query to guide their judgments. 
For the heatmap-based setting, only the chain-of-thought \cite{chain-of-thought} query is used.
As shown in Tables \ref{table:main_table_modality} and \ref{table:main_table_class}, despite being supported by powerful MLLMs, these approaches still underperform MedVeriSeg. 
We attribute this advantage to the multi-source verification design of our method, which integrates both qualitative and quantitative evidence to enable a more comprehensive and accurate analysis.
We further observe that, with the original image as input, these methods often show substantially lower F1 scores than accuracy, suggesting imbalanced performance between positive and negative samples.
The results in Table \ref{table:main_table_class} further support this hypothesis, as these methods show markedly weaker recognition performance on positive samples than on negative ones.
This suggests that their high performance on negative samples may be largely attributable to a prediction bias toward negative decisions, rather than to robust visual reasoning.
In contrast, MedVeriSeg shows a more balanced performance across positive and negative samples, indicating that it effectively reasons over diverse evidence and reaches reliable results.

\begin{figure}[t]
    \centering
    \includegraphics[width=1\linewidth]{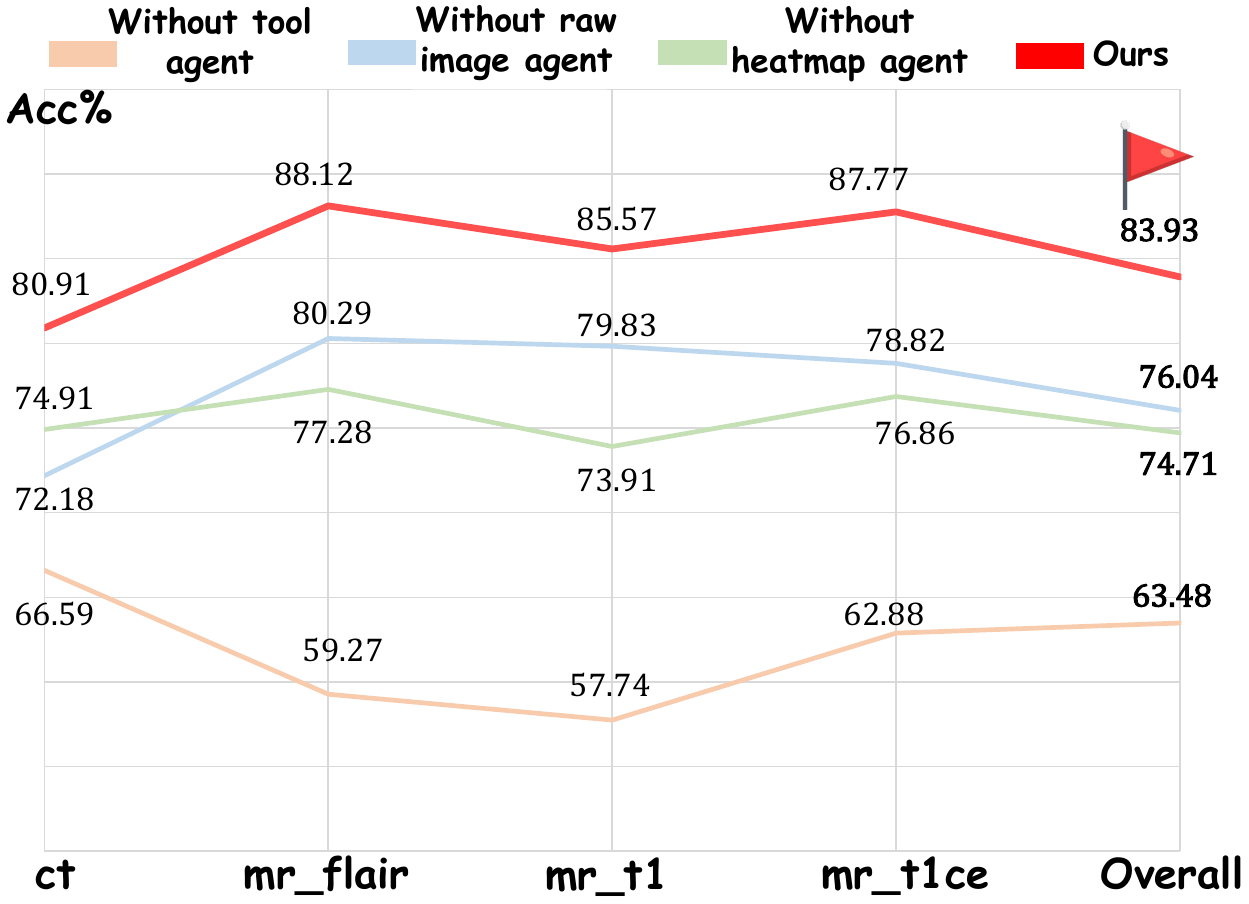}
    \vspace{-20pt}
    \caption{Ablation study of the three evidence-providing agents in MedVeriSeg. The values reported in the figure denote accuracy (Acc), reflecting the overall performance across both positive and negative samples.} 
    \label{fig:ab_3_agent}
\end{figure}

\begin{table}[t]
\centering
\setlength{\abovecaptionskip}{4pt}
\setlength{\belowcaptionskip}{4pt}
\caption{Ablation study of different score dimensions in the Similarity Response Quality Scoring Module.
\(\mathcal{S}\): Strength, \(\mathcal{C}\): Compactness, \(\mathcal{P}\): Purity.
}
\label{table:ab_scp}
\setlength{\tabcolsep}{4pt}
\footnotesize
\resizebox{\columnwidth}{!}{
\begin{tabular}{ccc|cc cc cc cc|cc}
\toprule
\multirow{3}{*}{\(\mathcal{S}\)}
& \multirow{3}{*}{\(\mathcal{C}\)}
& \multirow{3}{*}{\(\mathcal{P}\)}
& \multicolumn{10}{c}{\textbf{Easy Pos. + Easy Neg. Cases}} \\

\cmidrule(lr){4-13}

& & 
& \multicolumn{2}{c}{\textbf{ct}}
& \multicolumn{2}{c}{\textbf{mr\_flair}}
& \multicolumn{2}{c}{\textbf{mr\_t1}}
& \multicolumn{2}{c|}{\textbf{mr\_t1ce}}
& \multicolumn{2}{c}{\textbf{Overall}} \\

\cmidrule(lr){4-5}
\cmidrule(lr){6-7}
\cmidrule(lr){8-9}
\cmidrule(lr){10-11}
\cmidrule(lr){12-13}

& & 
& \textbf{Acc} & \textbf{F1}
& \textbf{Acc} & \textbf{F1}
& \textbf{Acc} & \textbf{F1}
& \textbf{Acc} & \textbf{F1}
& \textbf{Acc} & \textbf{F1} \\

\midrule
\Checkmark  &  &  
& 46.08 & 58.31
& 72.32 & 80.00
& 65.22 & 73.05
& 69.43 & 77.64
& 57.01 & 67.26\\

  & \Checkmark  &  
& 79.94 & 73.12
& 78.33 & 84.54
& 76.87 & 81.24
& 74.67 & 80.98
& 78.61 & 78.95 \\

  &  & \Checkmark  
& 71.25 & 70.35
& 67.62 & 79.23
& 66.96 & 76.43
& 69.65 & 79.59
& 69.74 & 74.97 \\

\Checkmark  & \Checkmark  &  
& 82.47 & 88.03
& 87.02 & 92.11
& 84.57 & 90.07
& 83.89 & 89.74
& 84.43 & 89.88 \\

\Checkmark  &  & \Checkmark  
& 65.41 & 76.64
& 77.92 & 87.22
& 74.03 & 83.82
& 77.71 & 86.42
& 71.49 & 81.93 \\

  & \Checkmark  & \Checkmark  
& 83.58 & 81.10
& 77.85 & 86.46
& 77.16 & 84.27
& 77.81 & 85.56
& 80.69 & 83.85 \\

\rowcolor{aliceblue!60}
\Checkmark  & \Checkmark  & \Checkmark  
& 83.45 & 89.44
& 86.35 & 92.43
& 85.58 & 91.49
& 85.23 & 91.31
& 84.93 & 90.97 \\

\bottomrule
\end{tabular}
}
\end{table}

\begin{figure*}[t]
    \centering
    \includegraphics[width=\textwidth]{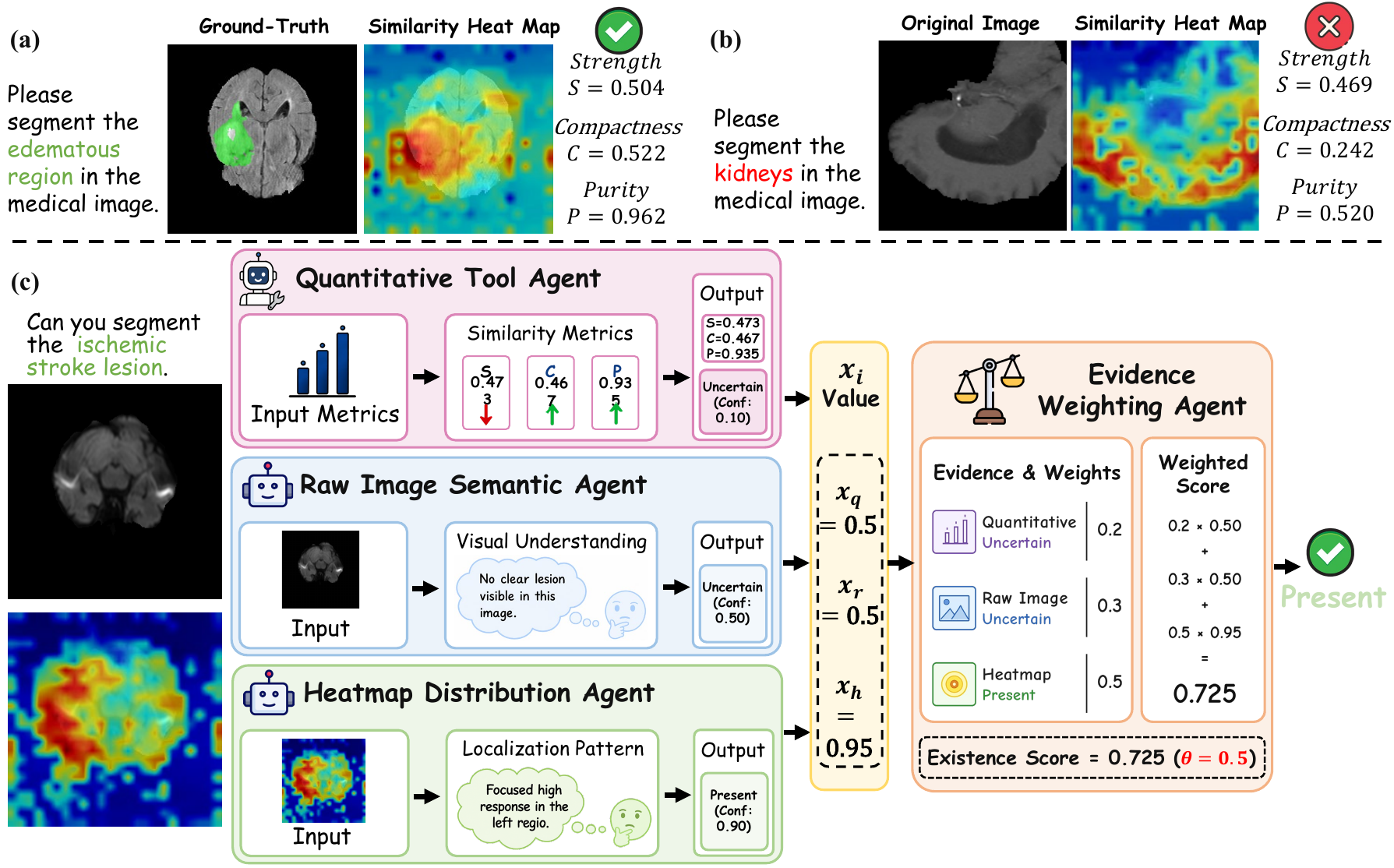}
    \vspace{-20pt}
    \caption{
Qualitative analysis and case studies of MedVeriSeg. (a) Easy positive cases, (b) easy negative cases, and (c) ambiguous cases.
    }
    \label{fig:case_study}
\end{figure*}

\subsection{Ablation Study}
\subsubsection{Choices of Agent MLLM Backbones}
We investigate the impact of different choices of agent MLLM backbones in the multi-agent module on the overall performance. As shown in Table \ref{table:ab_agent_mllm}, MedVeriSeg achieves the best performance when Gemini-2.5-Pro is used as the agent MLLM backbone.
However, it is not open-source. To ensure the reproducibility of our method, we ultimately adopt Qwen-3-VL, as it achieves the best performance among the open-source models.

\subsubsection{Choice of Metrics in SRQS}
In the Similarity Response Quality Scoring Module, we compute three metrics as quantitative evidence for the Quantitative Tool Agent. Table~\ref{table:ab_scp} summarizes the accuracy and F1 of different metric-selection strategies in identifying easy positive and negative cases, rather than reporting the overall accuracy. 
As shown in the table, using $S$, $C$, and $P$ jointly as quantitative evidence achieves the best performance. This result further supports the choice of metrics in our quantitative evidence design.

\subsubsection{Contribution of the Three Evidence-Providing Agents}

We investigate the relative contributions of the Quantitative Tool Agent, the Raw Image Semantic Agent, and the Heatmap Distribution Agent to the final results, thereby validating the rationality of our multi-agent module.
As shown in Figure \ref{fig:ab_3_agent}, removing any individual agent leads to varying degrees of performance degradation, demonstrating the effectiveness of all three agents. 
Notably, removing the Quantitative Tool Agent results in a substantial decline in performance, which further highlights the important role and value of the SRQS module in MedVeriSeg.

\begin{figure*}[t]
    \centering
    \includegraphics[width=\textwidth]{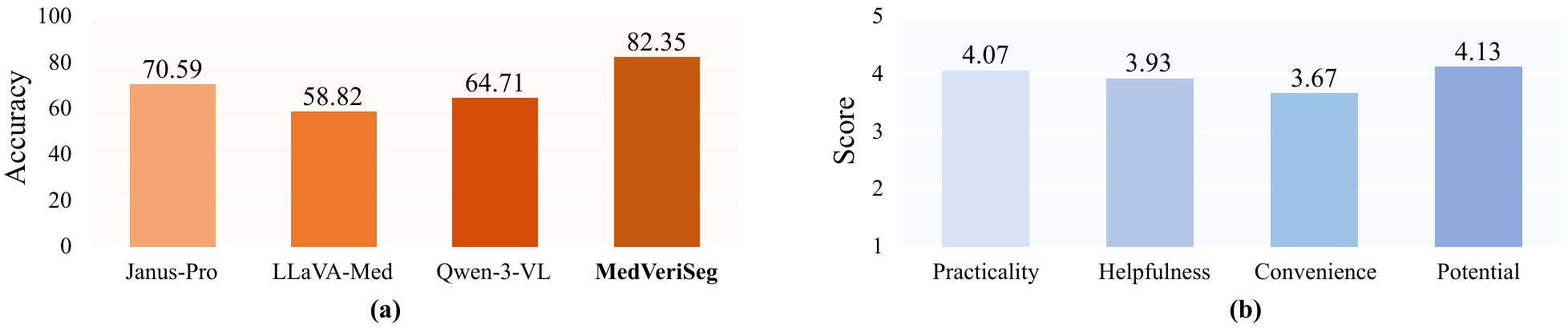}
    \vspace{-20pt}
    \caption{
Human evaluation results in terms of (a) human-query identification performance (Accuracy) and (b) practical utility assessment.
In figure (b), scores are rated on a 1–5 scale, with higher scores indicating better performance.
    }
    \label{fig:human_eval}
\end{figure*}

\subsubsection{Role of Quantitative Weighted Evidence Fusion}
At the backend of the LMAV module, we do not allow the Evidence Weighting Agent to directly output the final decision. 
Instead, the agent assigns scores to different evidence, which are then integrated through weighted evidence fusion to obtain the final score. 
The results in Table \ref{table:ab_score_fusion} demonstrate that, compared with allowing the Evidence Weighting Agent to directly produce the final result, this design achieves greater stability and better overall performance.

\subsection{Extension to Other Models}
We investigate the performance of MedVeriSeg when applied as a plug-in module to other models.
The results in Table \ref{table:plugin_generalization} show that MedVeriSeg remains effective on other LISA-like medical segmentation models, demonstrating that our method can serve as a plug-and-play module and highlighting its practical utility.
It is worth noting that the MedVeriSeg framework is not only effective for medical LISA-like segmentation models, but also remains effective on the general model, i.e., LISA \cite{lisa}, which further demonstrates its generalizability.

\begin{table}[t]
\centering
\setlength{\belowcaptionskip}{4pt}
\caption{Effect of quantitative weighted evidence fusion in the LMAV module. 
``EWA Direct Decision'' denotes the Evidence Weighting Agent directly makes the final prediction.
}
\label{table:ab_score_fusion}
\small
\setlength{\tabcolsep}{3pt}
\resizebox{\columnwidth}{!}{
\begin{tabular}{l|cccc}
\toprule
\textbf{Setting} & \textbf{Accuracy} & \textbf{F1} & \textbf{Precision} & \textbf{Recall}  \\

\midrule

EWA Direct Decision     &79.01 &77.50 &83.60  &72.23  \\
\rowcolor{aliceblue!60}
Ours    &83.93 &84.84 &80.30  &89.95    \\
\bottomrule
\end{tabular}
}
\end{table}

\begin{table}[t]
\centering
\setlength{\abovecaptionskip}{4pt}
\setlength{\belowcaptionskip}{4pt}
\caption{
Performance of MedVeriSeg as a plug-in module for other LISA-like segmentation models.
}
\label{table:plugin_generalization}
\setlength{\tabcolsep}{6.5pt}
\renewcommand{\arraystretch}{1.12}
\small
\resizebox{\columnwidth}{!}{
\begin{tabular}{llcccc}
\toprule
\textbf{Baseline} 
& \textbf{Setting}
& \textbf{Positive Acc.}
& \textbf{Negative Acc.}
& \textbf{Average Acc.}
& \textbf{Neg. Gain} \\
\midrule
\rowcolor{green!20}
\multicolumn{6}{c}{\textit{\textbf{Medical Model (test on MedVeriSeg-Bench)}}} \\
MediSee \cite{tong2025medisee}
& w/o plug-in 
& \textbf{100.00} & 0.00 & 50.00 & -- \\
\rowcolor{aliceblue!60}
MediSee
& + Ours 
& 89.95 & 77.89 & \textbf{83.92} & \textbf{+ 77.89} \\

MedPLIB \cite{huang2025towards_MedPLIB}
& w/o plug-in 
& \textbf{100.00} & 0.00 & 50.00 & -- \\
\rowcolor{aliceblue!60}
MedPLIB
& + Ours 
&83.90  &72.24  & \textbf{78.07} & \textbf{+ 72.24} \\

\midrule
\rowcolor{blue!20}
\multicolumn{6}{c}{\textit{\textbf{General Model (test on 175 natural images)}}} \\

LISA \cite{lisa}
& w/o plug-in 
& \textbf{100.00} & 0.00 & 50.00 & -- \\
\rowcolor{aliceblue!60}
LISA
& + Ours 
&94.12  &95.56  & \textbf{94.84} & \textbf{+ 95.56} \\

\bottomrule
\end{tabular}
}
\end{table}

\subsection{Qualitative Analysis and Case Study}

Figure \ref{fig:case_study} visualizes three types of cases classified by the Dynamic Router in the LMAV module: easy positive cases, easy negative cases, and ambiguous cases.
It can be observed that the Dynamic Router is able to distinguish relatively straightforward samples. 
For ambiguous cases, MedVeriSeg can still arrive at the correct conclusion through comprehensive and rigorous analysis.

\subsection{Human Evaluation}
We conduct a human evaluation \cite{vasey2022reporting} to demonstrate the superior performance and practical utility of MedVeriSeg. 
Specifically, users are asked to provide segmentation instructions for 102 medical images, including both true and false queries. 
As shown in Figure \ref{fig:human_eval} (a), MedVeriSeg achieves the best identification performance.
Furthermore, as illustrated in Figure \ref{fig:human_eval} (b), we recruit 15 users (8 medical students and 7 clinicians) to assess the practical value of MedVeriSeg from four different perspectives using a 1--5 rating scale. 
The results demonstrate the real-world utility of MedVeriSeg in supporting LISA-like medical segmentation models.

\section{Limitations}\label{sec:limitation}
Although MedVeriSeg achieves effective query verification without additional training, it still has several limitations. First, the dynamic routing strategy improves efficiency by directly handling easy positive and easy negative cases according to the SRQS scores. 
However, if an easy-routed case is incorrectly classified at this stage, the subsequent visual verification agents will not be activated, leaving limited opportunity for correction. 
Second, although Table \ref{table:plugin_generalization} shows that MedVeriSeg can serve as a general plug-in for different LISA-like segmentation models, its performance may still fluctuate with the capability of the underlying segmentation model. 
This is because the quality of the original model directly affects the representation of the \texttt{[SEG]} token and, consequently, the quality of the similarity matrix. 
Such variations may further influence the judgments of the Quantitative Tool Agent and the Heatmap Distribution Agent. 
Although the proposed multi-source evidence integration design can alleviate this issue to some extent, residual perturbations caused by different backbone models still remain.

\section{Conclusion}\label{sec:conclusion}
In this work, we address hallucinated segmentation in LISA-like MLLM-based medical image segmentation, where models may generate masks even when the queried target is absent from the image. To tackle this issue, we propose \textbf{MedVeriSeg}, a training-free query verification framework that combines \texttt{[SEG]} token response quality scoring with lightweight routed multi-agent verification to assess the validity of segmentation queries. 
We further construct \textbf{MedVeriSeg-Bench} to support systematic evaluation of query verification in medical image segmentation. 
Experimental results demonstrate that MedVeriSeg effectively reduces hallucinated segmentation while largely preserving the original segmentation capability of existing LISA-like medical segmentation models, highlighting its flexibility and practical value for reliable interactive medical image segmentation.



\section*{Acknowledgment}
This work was supported by National Natural Science Foundation of China under Grant No.62501532; Zhejiang Provincial Natural Science Foundation of China under Grant No.LQN26F020039; the Fundamental Research Funds of Zhejiang Sci-Tech University under Grant No.26232179-Y.

\ifCLASSOPTIONcaptionsoff
  \newpage
\fi



%



\bibliographystyle{IEEEtran}
\bibliography{ref}

@String(AAAI  = {AAAI})

@article{sam_med_2d_20m,
  title={Sa-med2d-20m dataset: Segment anything in 2d medical imaging with 20 million masks},
  author={Ye, Jin and Cheng, Junlong and Chen, Jianpin and Deng, Zhongying and Li, Tianbin and Wang, Haoyu and Su, Yanzhou and Huang, Ziyan and Chen, Jilong and Jiang, Lei and others},
  journal={arXiv preprint arXiv:2311.11969},
  year={2023}
}

@article{medsam_model,
  title={Segment anything in medical images},
  author={Ma, Jun and He, Yuting and Li, Feifei and Han, Lin and You, Chenyu and Wang, Bo},
  journal={Nature communications},
  volume={15},
  number={1},
  pages={654},
  year={2024},
  publisher={Nature Publishing Group UK London}
}

@inproceedings{lisa,
  title={Lisa: Reasoning segmentation via large language model},
  author={Lai, Xin and Tian, Zhuotao and Chen, Yukang and Li, Yanwei and Yuan, Yuhui and Liu, Shu and Jia, Jiaya},
  booktitle={Proceedings of the IEEE/CVF conference on computer vision and pattern recognition},
  pages={9579--9589},
  year={2024}
}

@article{lisa++,
  title={Lisa++: An improved baseline for reasoning segmentation with large language model},
  author={Yang, Senqiao and Qu, Tianyuan and Lai, Xin and Tian, Zhuotao and Peng, Bohao and Liu, Shu and Jia, Jiaya},
  journal={arXiv preprint arXiv:2312.17240},
  year={2023}
}

@inproceedings{lisa-READ,
  title={Reasoning to attend: Try to understand how< SEG> token works},
  author={Qian, Rui and Yin, Xin and Dou, Dejing},
  booktitle={Proceedings of the Computer Vision and Pattern Recognition Conference},
  pages={24722--24731},
  year={2025}
}

@inproceedings{glamm-cvpr2024,
  title={Glamm: Pixel grounding large multimodal model},
  author={Rasheed, Hanoona and Maaz, Muhammad and Shaji, Sahal and Shaker, Abdelrahman and Khan, Salman and Cholakkal, Hisham and Anwer, Rao M and Xing, Eric and Yang, Ming-Hsuan and Khan, Fahad S},
  booktitle={Proceedings of the IEEE/CVF Conference on Computer Vision and Pattern Recognition},
  pages={13009--13018},
  year={2024}
}

@inproceedings{perceptiongpt-cvpr2024,
  title={Perceptiongpt: Effectively fusing visual perception into llm},
  author={Pi, Renjie and Yao, Lewei and Gao, Jiahui and Zhang, Jipeng and Zhang, Tong},
  booktitle={Proceedings of the IEEE/CVF conference on computer vision and pattern recognition},
  pages={27124--27133},
  year={2024}
}

@article{multi-round-seg,
  title={Segllm: Multi-round reasoning segmentation},
  author={Wang, XuDong and Zhang, Shaolun and Li, Shufan and Kallidromitis, Konstantinos and Li, Kehan and Kato, Yusuke and Kozuka, Kazuki and Darrell, Trevor},
  journal={arXiv preprint arXiv:2410.18923},
  year={2024}
}

@article{llava,
  title={Visual instruction tuning},
  author={Liu, Haotian and Li, Chunyuan and Wu, Qingyang and Lee, Yong Jae},
  journal={Advances in neural information processing systems},
  volume={36},
  pages={34892--34916},
  year={2023}
}

@article{llava-med,
  title={Llava-med: Training a large language-and-vision assistant for biomedicine in one day},
  author={Li, Chunyuan and Wong, Cliff and Zhang, Sheng and Usuyama, Naoto and Liu, Haotian and Yang, Jianwei and Naumann, Tristan and Poon, Hoifung and Gao, Jianfeng},
  journal={Advances in Neural Information Processing Systems},
  volume={36},
  pages={28541--28564},
  year={2023}
}

@article{nature-llm-seg,
  title={LLM-driven multimodal target volume contouring in radiation oncology},
  author={Oh, Yujin and Park, Sangjoon and Byun, Hwa Kyung and Cho, Yeona and Lee, Ik Jae and Kim, Jin Sung and Ye, Jong Chul},
  journal={Nature Communications},
  volume={15},
  number={1},
  pages={9186},
  year={2024},
  publisher={Nature Publishing Group UK London}
}

@article{biomed-parse,
  title={A foundation model for joint segmentation, detection and recognition of biomedical objects across nine modalities},
  author={Zhao, Theodore and Gu, Yu and Yang, Jianwei and Usuyama, Naoto and Lee, Ho Hin and Kiblawi, Sid and Naumann, Tristan and Gao, Jianfeng and Crabtree, Angela and Abel, Jacob and others},
  journal={Nature methods},
  volume={22},
  number={1},
  pages={166--176},
  year={2025},
  publisher={Nature Publishing Group US New York}
}

@article{sam-med-2d-model,
  title={Sam-med2d},
  author={Cheng, Junlong and Ye, Jin and Deng, Zhongying and Chen, Jianpin and Li, Tianbin and Wang, Haoyu and Su, Yanzhou and Huang, Ziyan and Chen, Jilong and Jiang, Lei and others},
  journal={arXiv preprint arXiv:2308.16184},
  year={2023}
}

@inproceedings{imis_model,
  title={Interactive medical image segmentation: A benchmark dataset and baseline},
  author={Cheng, Junlong and Fu, Bin and Ye, Jin and Wang, Guoan and Li, Tianbin and Wang, Haoyu and Li, Ruoyu and Yao, He and Cheng, Junren and Li, JingWen and others},
  booktitle={Proceedings of the Computer Vision and Pattern Recognition Conference},
  pages={20841--20851},
  year={2025}
}

@article{gpt-4,
  title={Gpt-4 technical report},
  author={Achiam, Josh and Adler, Steven and Agarwal, Sandhini and Ahmad, Lama and Akkaya, Ilge and Aleman, Florencia Leoni and Almeida, Diogo and Altenschmidt, Janko and Altman, Sam and Anadkat, Shyamal and others},
  journal={arXiv preprint arXiv:2303.08774},
  year={2023}
}

@article{gpt-5,
  title={Openai gpt-5 system card},
  author={Singh, Aaditya and Fry, Adam and Perelman, Adam and Tart, Adam and Ganesh, Adi and El-Kishky, Ahmed and McLaughlin, Aidan and Low, Aiden and Ostrow, AJ and Ananthram, Akhila and others},
  journal={arXiv preprint arXiv:2601.03267},
  year={2025}
}

@article{deepseek-janus,
  title={Janus-pro: Unified multimodal understanding and generation with data and model scaling},
  author={Chen, Xiaokang and Wu, Zhiyu and Liu, Xingchao and Pan, Zizheng and Liu, Wen and Xie, Zhenda and Yu, Xingkai and Ruan, Chong},
  journal={arXiv preprint arXiv:2501.17811},
  year={2025}
}

@article{instruction-med-seg-onesentence-1,
  title={Medical image segmentation using deep learning: A survey},
  author={Wang, Risheng and Lei, Tao and Cui, Ruixia and Zhang, Bingtao and Meng, Hongying and Nandi, Asoke K},
  journal={IET image processing},
  volume={16},
  number={5},
  pages={1243--1267},
  year={2022},
  publisher={Wiley Online Library}
}

@article{chain-of-thought,
  title={Chain-of-thought prompting elicits reasoning in large language models},
  author={Wei, Jason and Wang, Xuezhi and Schuurmans, Dale and Bosma, Maarten and Xia, Fei and Chi, Ed and Le, Quoc V and Zhou, Denny and others},
  journal={Advances in neural information processing systems},
  volume={35},
  pages={24824--24837},
  year={2022}
}

@article{intro_first_seg-2,
  title={A review of medical image segmentation algorithms.},
  author={Ramesh, KKD and Kumar, G Kiran and Swapna, K and Datta, Debabrata and Rajest, S Suman},
  journal={EAI Endorsed Transactions on Pervasive Health \& Technology},
  volume={7},
  number={27},
  year={2021}
}

@article{intro_first_seg-3,
  title={Medical image segmentation methods, algorithms, and applications},
  author={Norouzi, Alireza and Rahim, Mohd Shafry Mohd and Altameem, Ayman and Saba, Tanzila and Rad, Abdolvahab Ehsani and Rehman, Amjad and Uddin, Mueen},
  journal={IETE Technical Review},
  volume={31},
  number={3},
  pages={199--213},
  year={2014},
  publisher={Taylor \& Francis}
}

@inproceedings{huang2025towards_MedPLIB,
  title={Towards a multimodal large language model with pixel-level insight for biomedicine},
  author={Huang, Xiaoshuang and Shen, Lingdong and Liu, Jia and Shang, Fangxin and Li, Hongxiang and Huang, Haifeng and Yang, Yehui},
  booktitle={Proceedings of the AAAI Conference on Artificial Intelligence},
  volume={39},
  number={4},
  pages={3779--3787},
  year={2025}
}

@article{yan2025medreasoner,
  title={Medreasoner: Reinforcement learning drives reasoning grounding from clinical thought to pixel-level precision},
  author={Yan, Zhonghao and Diao, Muxi and Yang, Yuxuan and Jing, Ruoyan and Xu, Jiayuan and Zhang, Kaizhou and Yang, Lele and Liu, Yanxi and Liang, Kongming and Ma, Zhanyu},
  journal={arXiv preprint arXiv:2508.08177},
  year={2025}
}

@inproceedings{tong2025medisee,
  title={Medisee: Reasoning-based pixel-level perception in medical images},
  author={Tong, Qinyue and Lu, Ziqian and Liu, Jun and Zheng, Yangming and Lu, Zhe-Ming},
  booktitle={Proceedings of the 33rd ACM International Conference on Multimedia},
  pages={2742--2751},
  year={2025}
}

@article{bai2023qwen,
  title={Qwen technical report},
  author={Bai, Jinze and Bai, Shuai and Chu, Yunfei and Cui, Zeyu and Dang, Kai and Deng, Xiaodong and Fan, Yang and Ge, Wenbin and Han, Yu and Huang, Fei and others},
  journal={arXiv preprint arXiv:2309.16609},
  year={2023}
}

@article{huang2025medsegR,
  title={Medseg-r: Reasoning segmentation in medical images with multimodal large language models},
  author={Huang, Yu and Peng, Zelin and Zhao, Yichen and Yang, Piao and Yang, Xiaokang and Shen, Wei},
  journal={arXiv preprint arXiv:2506.10465},
  year={2025}
}

@article{team2023gemini,
  title={Gemini: a family of highly capable multimodal models},
  author={Team, Gemini and Anil, Rohan and Borgeaud, Sebastian and Alayrac, Jean-Baptiste and Yu, Jiahui and Soricut, Radu and Schalkwyk, Johan and Dai, Andrew M and Hauth, Anja and Millican, Katie and others},
  journal={arXiv preprint arXiv:2312.11805},
  year={2023}
}

@article{tong2025mediround,
  title={MediRound: Multi-Round Entity-Level Reasoning Segmentation in Medical Images},
  author={Tong, Qinyue and Lu, Ziqian and Liu, Jun and Zuo, Rui and Lu, Zheming and Jin, Yueming},
  journal={arXiv preprint arXiv:2511.12110},
  year={2025}
}

@inproceedings{gsva_existing_work_general,
  title={Gsva: Generalized segmentation via multimodal large language models},
  author={Xia, Zhuofan and Han, Dongchen and Han, Yizeng and Pan, Xuran and Song, Shiji and Huang, Gao},
  booktitle={Proceedings of the IEEE/CVF Conference on Computer Vision and Pattern Recognition},
  pages={3858--3869},
  year={2024}
}

@inproceedings{liu2023gres_existing_work_general_2,
  title={Gres: Generalized referring expression segmentation},
  author={Liu, Chang and Ding, Henghui and Jiang, Xudong},
  booktitle={Proceedings of the IEEE/CVF conference on computer vision and pattern recognition},
  pages={23592--23601},
  year={2023}
}

@inproceedings{see_say_segment,
  title={See say and segment: Teaching lmms to overcome false premises},
  author={Wu, Tsung-Han and Biamby, Giscard and Chan, David and Dunlap, Lisa and Gupta, Ritwik and Wang, Xudong and Gonzalez, Joseph E and Darrell, Trevor},
  booktitle={Proceedings of the IEEE/CVF Conference on Computer Vision and Pattern Recognition},
  pages={13459--13469},
  year={2024}
}

@inproceedings{tang2025ltse,
  title={LTSE: Language-Guided Tissue Referring Segmentation in Pathology Images with Adaptive Expert Mixture},
  author={Tang, Jiao and Qian, Bo and Wan, Peng and Shao, Wei and Zhang, Daoqiang},
  booktitle={International Conference on Medical Image Computing and Computer-Assisted Intervention},
  pages={405--415},
  year={2025},
  organization={Springer}
}

@inproceedings{choi2026instruction,
  title={Instruction-Guided Lesion Segmentation for Chest X-rays with Automatically Generated Large-Scale Dataset},
  author={Choi, Geon and Yoon, Hangyul and Shin, Hyunju and Park, Hyunki and Seo, Sang Hoon and Yang, Eunho and Choi, Edward},
  booktitle={Proceedings of the IEEE/CVF Conference on Computer Vision and Pattern Recognition},
  pages={1482--1492},
  year={2026}
}

@article{team2025kimi,
  title={Kimi k2: Open agentic intelligence},
  author={Team, Kimi and Bai, Yifan and Bao, Yiping and Charles, Y and Chen, Cheng and Chen, Guanduo and Chen, Haiting and Chen, Huarong and Chen, Jiahao and Chen, Ningxin and others},
  journal={arXiv preprint arXiv:2507.20534},
  year={2025}
}

@article{csvt_medical_seg_1,
  title={Uncertainty-aware hierarchical aggregation network for medical image segmentation},
  author={Zhou, Tao and Zhou, Yi and Li, Guangyu and Chen, Geng and Shen, Jianbing},
  journal={IEEE Transactions on Circuits and Systems for Video Technology},
  volume={34},
  number={8},
  pages={7440--7453},
  year={2024},
  publisher={IEEE}
}

@article{csvt_medical_seg_2,
  title={Cnn-transformer rectified collaborative learning for medical image segmentation},
  author={Wu, Lanhu and Zhang, Miao and Piao, Yongri and Yao, Zhenyan and Sun, Weibing and Tian, Feng and Lu, Huchuan},
  journal={IEEE Transactions on Circuits and Systems for Video Technology},
  volume={35},
  number={5},
  pages={4072--4086},
  year={2024},
  publisher={IEEE}
}

@article{csvt_medical_seg_3,
  title={Ultra-lightweight network for medical image segmentation inspired by bio-visual interaction},
  author={Cai, Zhefei and Fan, Yingle and Zhu, Minwei and Fang, Tao},
  journal={IEEE Transactions on Circuits and Systems for Video Technology},
  volume={35},
  number={4},
  pages={3486--3497},
  year={2024},
  publisher={IEEE}
}

@article{csvt_medical_seg_4,
  title={Frequency-aware interaction network for ultrasound image segmentation},
  author={Wang, Dongfang and Zhou, Tao and Zhang, Yizhe and Gao, Shangbing and Yang, Jian},
  journal={IEEE Transactions on Circuits and Systems for Video Technology},
  volume={35},
  number={7},
  pages={7020--7032},
  year={2024},
  publisher={IEEE}
}

@article{tong2025chat,
  title={Chat-MedGen: omni-adaptation of multi-modal large language models for diverse biomedical tasks},
  author={Tong, Qinyue and Lu, Ziqian and Lu, Zheming and Yu, Yunlong and Liu, Jun and Zheng, Yangming},
  journal={Knowledge-Based Systems},
  pages={115007},
  year={2025},
  publisher={Elsevier}
}

@article{csvt_mllm_1,
  title={Surveillance video-and-language understanding: from small to large multimodal models},
  author={Yuan, Tongtong and Zhang, Xuange and Liu, Bo and Liu, Kun and Jin, Jian and Jiao, Zhenzhen},
  journal={IEEE Transactions on Circuits and Systems for Video Technology},
  volume={35},
  number={1},
  pages={300--314},
  year={2024},
  publisher={IEEE}
}

@article{csvt_mllm_2,
  title={Mg-llava: Towards multi-granularity visual instruction tuning},
  author={Zhao, Xiangyu and Li, Xiangtai and Duan, Haodong and Huang, Haian and Li, Yining and Chen, Kai and Yang, Hua},
  journal={IEEE Transactions on Circuits and Systems for Video Technology},
  year={2025},
  publisher={IEEE}
}

@article{fawcett2006introduction,
  title={An introduction to ROC analysis},
  author={Fawcett, Tom},
  journal={Pattern recognition letters},
  volume={27},
  number={8},
  pages={861--874},
  year={2006},
  publisher={Elsevier}
}

@article{vasey2022reporting,
  title={Reporting guideline for the early stage clinical evaluation of decision support systems driven by artificial intelligence: DECIDE-AI},
  author={Vasey, Baptiste and Nagendran, Myura and Campbell, Bruce and Clifton, David A and Collins, Gary S and Denaxas, Spiros and Denniston, Alastair K and Faes, Livia and Geerts, Bart and Ibrahim, Mudathir and others},
  journal={bmj},
  volume={377},
  year={2022},
  publisher={British Medical Journal Publishing Group}
}

@article{zhang2023multimodalcot,
  title={Multimodal chain-of-thought reasoning in language models},
  author={Zhang, Zhuosheng and Zhang, Aston and Li, Mu and Zhao, Hai and Karypis, George and Smola, Alex},
  journal={arXiv preprint arXiv:2302.00923},
  year={2023}
}

@inproceedings{suris2023vipergpt,
  title={Vipergpt: Visual inference via python execution for reasoning},
  author={Sur{\'\i}s, D{\'\i}dac and Menon, Sachit and Vondrick, Carl},
  booktitle={Proceedings of the IEEE/CVF international conference on computer vision},
  pages={11888--11898},
  year={2023}
}

@article{yang2023mm,
  title={Mm-react: Prompting chatgpt for multimodal reasoning and action},
  author={Yang, Zhengyuan and Li, Linjie and Wang, Jianfeng and Lin, Kevin and Azarnasab, Ehsan and Ahmed, Faisal and Liu, Zicheng and Liu, Ce and Zeng, Michael and Wang, Lijuan},
  journal={arXiv preprint arXiv:2303.11381},
  year={2023}
}

\end{document}